\documentclass[letterpaper,journal]{IEEEtran}

\usepackage{amsmath,amssymb,amsfonts}
\usepackage{graphicx}
\usepackage{booktabs}
\usepackage{multirow}
\usepackage{adjustbox}
\usepackage{tabularx}
\usepackage{array}
\usepackage{threeparttable}
\usepackage{float}
\usepackage{placeins}
\usepackage[ruled,vlined,linesnumbered]{algorithm2e}
\usepackage{cite}
\usepackage{url}
\usepackage{xcolor}
\usepackage[hidelinks]{hyperref}
\hypersetup{
    colorlinks=true,
    citecolor=blue,
    linkcolor=blue,
    urlcolor=blue
}

\usepackage{etoolbox}

\AtBeginEnvironment{thebibliography}{%
  \setlength{\itemsep}{0pt}%
  \setlength{\parskip}{0pt}%
  \setlength{\parsep}{0pt}%
}

\SetKwFor{ForAll}{for all}{do}{end}
\SetKw{Return}{return}
\SetKwInput{KwParams}{Parameters}
\SetInd{0.5em}{1.2em}
\SetAlgoNlRelativeSize{-1}
\DontPrintSemicolon

\begin{document}

\title{Confidence-Guided Pseudo-Label Refinement for Unsupervised Cross-Domain Person Re-Identification}

\author{Sundas Iqbal, Danish Ali, Qing Tian, Yuhui Zheng, Jianping Gou, and Weihua Ou%

\thanks{This work was supported by the National Natural Science Foundation of China under Grant 62176128, the Basic Research Program of Jiangsu under Grant BK20231143, the Open Project of Key Laboratory of Tibetan Information Processing of the Ministry of Education under Grant QHSF-CS-2609, the Fundamental Research Funds for the Central Universities No. NJ2023032, as well as the 333 High-Level Talent Project of Jiangsu Province.}%

\thanks{Sundas Iqbal is with the School of Software, Nanjing University of Information Science and Technology, Nanjing 210044, China (e-mail: sundasiqbal058@gmail.com).}%

\thanks{Danish Ali is with the School of Computer Science, Wuhan University, Wuhan 430072, China (e-mail: danishalikhan545@gmail.com).}%

\thanks{Qing Tian is with the School of Computer Science, Nanjing University of Information Science and Technology, Nanjing 210044, China; the Wuxi Institute of Technology, Nanjing University of Information Science and Technology, Wuxi 214000, China; and the Key Laboratory of Tibetan Information Processing of the Ministry of Education, Qinghai Normal University, Xining 810008, China (corresponding author, e-mail: tianqing@nuist.edu.cn).}%

\thanks{Yuhui Zheng is with the College of Computer, Qinghai Normal University, Xining 810016, China (e-mail: zhengyh@vip.126.com).}%

\thanks{Jianping Gou is with the School of Computer and Information Science, Southwest University, Chongqing 400715, China (e-mail: cherish.gjp@gmail.com).}%

\thanks{Weihua Ou is with the School of Big Data and Computer Science, Guizhou Normal University, Guiyang 550025, China (e-mail: ouweihua@gznu.edu.cn).}%

}

\maketitle

\begin{abstract}
Unsupervised cross-domain person re-identification remains challenging because domain shifts generate noisy pseudo labels that are reinforced during iterative adaptation. Existing clustering-based methods often rely on limited reliability cues or global confidence criteria and may impose hard identity supervision on ambiguous target samples. To address these limitations, we propose CGPLR, a confidence-guided pseudo-label refinement framework for unsupervised domain-adaptive person re-identification. CGPLR estimates sample-level reliability by jointly modeling centroid compatibility, local neighborhood consistency, separation from the strongest competing cluster and temporal assignment stability. The resulting confidence distribution is used to derive cluster-adaptive upper and lower thresholds for a three-way selective decision. Reliable assignments are preserved, recoverable samples are reassigned only when an alternative cluster receives sufficient geometric and neighborhood support and highly ambiguous samples are deferred from hard identity supervision. The learning objective adapts to these decisions through confidence-weighted supervision for preserved samples, cautious corrected-label weighting for reassigned samples and label-free representation consistency for deferred samples. CGPLR achieves 76.4\% Rank-1 accuracy and 47.6\% mean average precision (mAP) on Market-1501$\rightarrow$MSMT17, and 95.1\% Rank-1 accuracy and 87.7\% mAP on MSMT17$\rightarrow$Market-1501, demonstrating effective reliability-aware selective adaptation. The code is available at \url{https://github.com/Sundas058/CGPLR}.
\end{abstract}

\begin{IEEEkeywords}
Unsupervised domain adaptation, person re-identification, pseudo-label refinement, confidence-aware learning.
\end{IEEEkeywords}

\section{Introduction}

\IEEEPARstart{P}{erson} re-identification (Re-ID) aims to match individuals across non-overlapping camera views and is widely used in intelligent surveillance and public security applications \cite{gao2024semantic,pandeeswari2025deep}. Although supervised deep learning has achieved substantial progress, it relies heavily on large-scale identity annotations that are costly and time-consuming to obtain \cite{ran2025camera,wang2026structured}. Unsupervised domain adaptation (UDA) addresses this limitation by transferring a model trained on a labeled source domain to an unlabeled target domain \cite{zhang2025heterogeneous,tian2025duel,Yang2026MEFAN}. Most UDA-based Re-ID methods therefore cluster target-domain features to generate pseudo labels for unsupervised adaptation \cite{li2026dynamic,tian2025unsupervised}.

However, clustering under domain shift inevitably produces incorrect assignments because of illumination, occlusion, viewpoint and background variations \cite{chen2026multigranularity,qin2026oneshot}, and these errors may be progressively reinforced during iterative adaptation \cite{yang2026attention,zheng2025threec}. Existing refinement and confidence-aware methods often rely on limited or static reliability cues. High compatibility with the assigned cluster may still be unreliable when a sample is also close to a competing cluster, while neighborhood agreement can be misleading when nearby assignments are noisy. Moreover, global confidence thresholds overlook variations in cluster density and compactness, and uncertain samples may be uniformly down-weighted or forcibly reassigned even when no alternative identity is sufficiently supported. More importantly, confidence is commonly used only as a loss weight rather than as a decision variable that determines both pseudo-label refinement and the form of supervision. Consequently, existing pipelines remain vulnerable to error propagation under severe domain shifts.

To address these limitations, we propose CGPLR, a confidence-guided pseudo-label refinement framework for unsupervised cross-domain person re-identification. CGPLR estimates sample reliability from centroid compatibility, neighborhood consistency, competing-cluster separation and temporal assignment stability. The resulting multi-evidence confidence is used to construct cluster-adaptive upper and lower thresholds for a three-way selective decision: reliable assignments are preserved, recoverable assignments are reassigned only when an alternative cluster receives sufficient geometric and neighborhood support and highly ambiguous samples are deferred from hard identity supervision. The learning objective is conditioned on these decisions with confidence-weighted supervision for preserved samples, cautious corrected-label weighting for reassigned samples and label-free representation consistency for deferred samples. Thus, reliability determines not only sample importance but also the type of supervision applied during adaptation, enabling recoverable errors to be corrected without forcing ambiguous samples into unreliable identity clusters. The main distinction of CGPLR is therefore the coupling of multi-evidence reliability estimation with cluster-adaptive label decisions and sample-specific learning behavior within a unified adaptation framework.

The main contributions of this work are summarized as follows:

\begin{itemize}

    \item We propose CGPLR, a reliability-aware framework for unsupervised cross-domain person re-identification that connects confidence estimation, selective pseudo-label refinement and target-domain optimization within a unified iterative adaptation process.

    \item We develop a multi-evidence confidence estimation strategy that evaluates each target sample using centroid compatibility, neighborhood consistency, competing-cluster separation and temporal assignment stability, providing a more complete measure of pseudo-label reliability.

    \item We introduce a cluster-adaptive selective refinement mechanism that preserves reliable assignments, reassigns recoverable samples only when an alternative cluster is consistently supported and defers genuinely ambiguous samples from hard pseudo-label supervision.

    \item We design a reliability-aware selective learning objective in which preserved, reassigned and deferred samples receive different forms of supervision including confidence-weighted classification, metric learning and label-free representation consistency.

    \item We conduct extensive experiments on multiple benchmark datasets and cross-domain transfer settings, supported by ablation, sensitivity, pseudo-label reliability and qualitative analyses to demonstrate the effectiveness and robustness of the proposed framework.

\end{itemize}

The remainder of this paper is organized as follows. Section~\ref{sec:related} reviews related work, Section~\ref{sec:method} presents CGPLR, Section~\ref{sec:exp} reports the experiments and Section~\ref{sec:conclusion} concludes the paper.

\section{Related Work}
\label{sec:related}

\subsection{Unsupervised Domain Adaptation for Person Re-ID}

Unsupervised domain adaptation (UDA) for person re-identification transfers identity-discriminative knowledge from a labeled source domain to an unlabeled target domain without target identity annotations \cite{iqbal2026cdmmd,tian2025cross,Zhang2026OpenBPR}. A dominant strategy follows iterative self-training, where target features are clustered to generate pseudo labels that are subsequently used to update the model \cite{tang2026unsupervised,yang2026multilevel}. Representative methods include Self-Paced Contrastive Learning (SPCL) \cite{ge2020self}, which combines hybrid memory with self-paced learning, and Cluster Contrast Learning (CCL) \cite{dai2022ccl}, which exploits cluster-level representations for pseudo-label supervision. Zhang et al.~\cite{zhang2024asymmetric} introduced asymmetric double-network mutual teaching, while Tan et al.~\cite{tan2026ahn} studied continual adaptation through affinity harmonization and Camera-Driven Representation Learning \cite{lee2023camera} explicitly modeled camera-induced variations. In the source-free setting, Tian et al.~\cite{tian2024dcl} proposed Dipolar Confidence Learning (DCL) to regulate target-domain adaptation without access to source data. Although these approaches improve transfer through contrastive learning, camera-aware modeling, representation alignment or confidence regulation, they remain dependent on the quality of target-domain clustering, allowing incorrect or unstable assignments to be reinforced during iterative adaptation, particularly under severe domain shifts.

\subsection{Pseudo-Label Learning for Unsupervised Person Re-ID}

Pseudo-label learning is central to unsupervised person Re-ID because it enables discriminative representation learning from unlabeled target data \cite{zhang2024asymmetric,Liu2026Confidence}. However, clustering-based pseudo labels are inherently noisy and may suffer from identity fragmentation, cluster contamination and boundary ambiguity. Tang et al.~\cite{tang2026unsupervised} proposed camera-aware multi-level refinement at cluster and instance levels, while Cho et al.~\cite{cho2022pplr} introduced Part-Based Pseudo Label Refinement (PPLR) to exploit complementary global and part-level representations. Zou et al.~\cite{zou2023dcmip} used discrepant and multi-instance proxies for clustered-sample supervision, Zhu et al.~\cite{zhu2024dkdmpl} employed multiview knowledge distillation to suppress pseudo-label noise, Peng et al.~\cite{peng2025cde} reduced camera deviation through CDE-Learning and Ding et al.~\cite{ding2026multi} combined multi-scale adaptive clustering with local-consistency constraints. Although these methods improve pseudo-label quality from different perspectives, they primarily focus on constructing or refining assignments and may still impose hard identity supervision on uncertain samples when neither the current nor an alternative assignment is sufficiently supported.

\subsection{Confidence and Reliability-Aware Learning}

Sample-level confidence provides an effective means of distinguishing reliable target instances from ambiguous or noisy pseudo-labeled samples \cite{wang2026temporal,tian2026progressive}. Miao et al.~\cite{miao2024confidence} proposed Confidence-Guided Centroids (CGC), which emphasizes high-confidence instances during centroid construction to reduce the influence of ambiguous boundary samples. Tian et al.~\cite{tian2024dcl} incorporated dipolar confidence into source-free adaptation, while Zheng et al.~\cite{zheng2025threec} combined confidence-guided clustering with contrastive learning for unsupervised person Re-ID. Reliability modeling has also been explored in specialized settings. Peng et al.~\cite{peng2026bi} proposed BIMM-Net for unsupervised visible--infrared person re-identification, using confidence-based matching to reduce noisy cross-modal associations. Complementary supervision has further been obtained through multi-view knowledge distillation \cite{zhu2024dkdmpl}, while denoising-oriented approaches suppress unreliable pseudo labels through feature denoising and teacher--student learning \cite{ding2026multi,tian2025part}.

\begin{table}[!t]
\centering
\caption{Comparison of representative confidence- and pseudo-label-refinement methods for unsupervised person re-identification.}
\label{tab:method_positioning}

\scriptsize
\setlength{\tabcolsep}{2.2pt}
\renewcommand{\arraystretch}{1.02}

\begin{tabularx}{\columnwidth}
{@{}>{\raggedright\arraybackslash}Xccccc@{}}
\toprule

Mechanism
& \shortstack{CGC\\\cite{miao2024confidence}}
& \shortstack{3C\\\cite{zheng2025threec}}
& \shortstack{PPLR\\\cite{cho2022pplr}}
& \shortstack{CMLR\\\cite{tang2026unsupervised}}
& \shortstack{\textbf{CGPLR}\\\textbf{(Ours)}} \\

\midrule

Centroid-based reliability
& $\checkmark$
& $\checkmark$
& --
& --
& $\checkmark$ \\

Local/neighborhood evidence
& --
& --
& $\checkmark$
& $\checkmark$
& $\checkmark$ \\

Explicit centroid margin
& --
& --
& --
& --
& $\checkmark$ \\

Temporal stability
& --
& --
& --
& --
& $\checkmark$ \\

Cluster-adaptive threshold
& --
& $\checkmark$
& --
& --
& $\checkmark$ \\

Evidence-gated hard reassignment
& --
& --
& --
& --
& $\checkmark$ \\

Selective deferral
& --
& --
& --
& --
& $\checkmark$ \\

Reliability-aware supervision
& $\checkmark$
& $\checkmark$
& $\checkmark$
& $\checkmark$
& $\checkmark$ \\

\bottomrule
\end{tabularx}

\end{table}

Existing confidence-aware approaches improve robustness from different perspectives. CGC \cite{miao2024confidence} emphasizes reliable samples during centroid construction, 3C \cite{zheng2025threec} incorporates confidence into clustering and contrastive learning, PPLR \cite{cho2022pplr} exploits agreement between global and part-level representations and CMLR \cite{tang2026unsupervised} performs camera-aware refinement at cluster and instance levels. However, these approaches do not jointly model whether a sample is simultaneously compatible with competing clusters, temporally stable across adaptation rounds and supported by a cluster-specific reliability distribution. CGPLR addresses these limitations by combining centroid compatibility, neighborhood consistency, competing-centroid separation and temporal assignment stability. The resulting confidence drives cluster-adaptive preserve, reassign or defer decisions and determines the form and strength of supervision applied to each target sample. As summarized in Table~\ref{tab:method_positioning}, the main distinction of CGPLR is the coupling of multi-evidence reliability estimation with cluster-adaptive pseudo-label decisions and reliability-conditioned learning within a unified adaptation framework.

\section{Proposed Method}
\label{sec:method}

\subsection{Problem Formulation}
\label{sec:problem_formulation}

Unsupervised domain adaptation (UDA) for person re-identification aims
to transfer identity-discriminative knowledge from a labeled source
domain to an unlabeled target domain under distribution shift. Let the
source domain be denoted as
$\mathcal{D}_s=\{(x_i^s,y_i^s)\}_{i=1}^{N_s}$, where $x_i^s$ is the
$i$-th source image and
$y_i^s\in\{1,\ldots,C_s\}$ is its identity label. The unlabeled target
domain is defined as
$\mathcal{D}_t=\{x_j^t\}_{j=1}^{N_t}$, where target identity annotations
are unavailable.

The objective is to learn a feature embedding function
$f_\theta:\mathcal{X}\rightarrow\mathbb{R}^{d}$ that preserves
identity-discriminative structure in the target domain. Since the
target identities are unknown, clustering-based pseudo-labeling is
used to approximate the latent target identity distribution.
Specifically, the pseudo label of target sample $x_j^t$ is generated
according to Eq.~\eqref{eq:pseudo_label}:
\begin{equation}
\label{eq:pseudo_label}
\hat{y}_j^t
=
\mathcal{C}
\left(
f_\theta(x_j^t)
\right),
\end{equation}
where $\mathcal{C}(\cdot)$ denotes a clustering operator such as
DBSCAN. The assignment produced by Eq.~\eqref{eq:pseudo_label}
provides an initial supervisory signal for target-domain adaptation.

However, due to domain discrepancy, the generated pseudo labels may
be incorrect. Conceptually, the correctness of a target pseudo label
is represented by the latent indicator
$\delta_j=\mathbb{I}(\hat{y}_j^t=y_j^t)$, where $y_j^t$ denotes the
unknown ground-truth identity. Because $\delta_j$ is unavailable
during adaptation, treating all pseudo-labeled samples equally may
introduce biased supervision and repeatedly reinforce incorrect
cluster assignments. To reduce the influence of unreliable pseudo labels, target-domain
adaptation is formulated as the confidence-weighted risk minimization
problem given in Eq.~\eqref{eq:weighted_risk}:
\begin{equation}
\label{eq:weighted_risk}
\mathcal{L}_t
=
\mathbb{E}_{x_j^t\sim\mathcal{D}_t}
\left[
w_j\,
\ell
\left(
f_\theta(x_j^t),
\hat{y}_j^t
\right)
\right],
\end{equation}
where $w_j\in[0,1]$ denotes the reliability weight of sample $x_j^t$ and serves as a surrogate for its unobservable pseudo-label
correctness probability,
$w_j\approx\mathbb{P}(\delta_j=1\mid x_j^t)$. As expressed in Eq.~\eqref{eq:weighted_risk}, reliable samples contribute more strongly
to target-domain learning, whereas uncertain assignments exert a smaller influence on optimization.

CGPLR follows a two-stage optimization procedure rather than jointly
optimizing the source- and target-domain objectives. The encoder is first
trained on the labeled source domain and is subsequently adapted to the
unlabeled target domain through reliability-aware supervision. The two
optimization stages are formulated in
Eq.~\eqref{eq:two_stage_objective}:
\begin{equation}
\label{eq:two_stage_objective}
\begin{aligned}
\theta^{(0)}
&=
\operatorname*{arg\,min}_{\theta}
\mathcal{L}_{s}(\theta),
\\
\theta^{*}
&=
\operatorname*{arg\,min}_{\theta}
\mathcal{L}_{t}(\theta),
\qquad
\theta_{\mathrm{init}}=\theta^{(0)}.
\end{aligned}
\end{equation}
where $\theta^{(0)}$ denotes the source-trained encoder parameters,
while $\theta^{*}$ denotes the parameters obtained after target-domain
adaptation. The source objective $\mathcal{L}_{s}$ first provides an
identity-discriminative initialization, after which the target objective
$\mathcal{L}_{t}$ adapts the encoder to the unlabeled target domain using
reliability-aware supervision. The target-domain adaptation stage follows the confidence-weighted risk formulation in Eq.~\eqref{eq:weighted_risk}. Therefore, the central
challenge is to estimate reliable sample weights $w_j$ and use them to regulate both pseudo-label refinement and target-domain optimization.
The following subsections describe how CGPLR addresses this challenge through multi-evidence confidence estimation, adaptive selective
refinement and reliability-aware learning.

\subsection{Framework Overview}
\label{subsec:framework_overview}

\begin{figure*}[!t]
\centering
    \includegraphics[
        width=0.80\textwidth
    ]{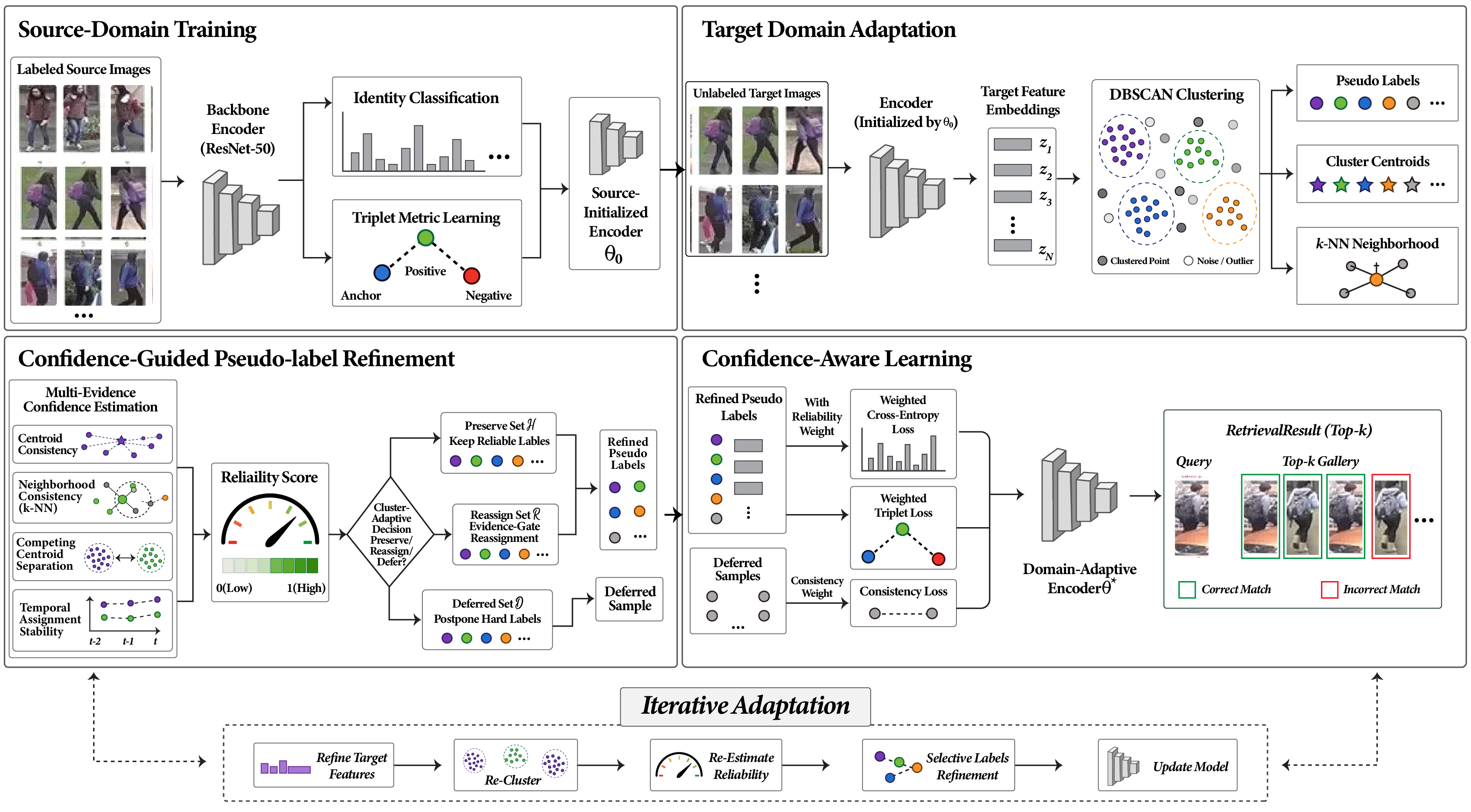}
    
    \caption{Overview of the proposed CGPLR framework for
    unsupervised cross-domain person re-identification. Target pseudo
    labels are evaluated using structural, competing-cluster and temporal
    evidence, followed by cluster-adaptive preserve, reassign or defer
    decisions and reliability-conditioned target-domain learning.}
    \label{fig:framework}
\end{figure*}

An overview of the proposed confidence-guided pseudo-label refinement framework is illustrated in Fig.~\ref{fig:framework}. CGPLR follows an iterative target-domain adaptation process that integrates pseudo-label generation, multi-evidence reliability estimation, selective refinement and reliability-aware learning. CGPLR estimates the reliability of each assignment and uses this information to determine both the refinement decision and the form of supervision applied during adaptation. Given a labeled source domain $\mathcal{D}_s$ and an unlabeled target domain $\mathcal{D}_t$, the encoder $f_\theta$ is first trained on the
source data and then used to extract target-domain features, which are clustered to generate initial pseudo labels. Sample reliability is assessed
using centroid compatibility, neighborhood consistency, competing-cluster separation and temporal assignment stability. Cluster-adaptive thresholds,
together with the evidence supporting alternative clusters, partition valid target samples into preserved, reassigned and deferred subsets. Reliable assignments are retained, uncertain assignments are corrected only when an alternative cluster is sufficiently supported and highly ambiguous samples are excluded from hard identity supervision. Preserved and reassigned samples are optimized through reliability-weighted classification and metric-learning losses, whereas deferred samples
contribute through label-free representation consistency. After each adaptation round, the updated encoder generates new target representations,
while the current pseudo labels and cluster centroids are retained as temporal information for the subsequent round. This process forms a
progressive refinement loop that connects clustering, selective correction and reliability-aware learning. The complete training procedure is summarized in Algorithm~\ref{alg:framework} presented at the end of this section.


\subsection{Source Domain Training}
\label{sec:source_training}

The source-domain training stage learns an identity-discriminative representation that provides a reliable initialization for subsequent
target-domain adaptation. Given the labeled source dataset $\mathcal{D}_s=\{(x_i^s,y_i^s)\}_{i=1}^{N_s}$, the feature encoder
$f_\theta$ is optimized under full supervision using identity classification and triplet-based metric learning. The classification objective encourages separation among source identities, whereas the triplet objective structures the embedding space by reducing intra-identity distances and increasing inter-identity separation. The corresponding classification loss, triplet loss and complete source-domain objective are formulated in
Eq.~\eqref{eq:source_training}:
\begin{equation}
\label{eq:source_training}
\begin{aligned}
\mathcal{L}_{\mathrm{ce}}
&=
-\sum_{i=1}^{N_s}
\log p_\theta\!\left(y_i^s \mid x_i^s\right),
\\[1mm]
\mathcal{L}_{\mathrm{tri}}
&=
\sum_{i=1}^{N_s}
\Big[
\left\|f_\theta(x_i^s)-f_\theta(x_i^{s+})\right\|_2^2
\\[-1mm]
&\qquad
-
\left\|f_\theta(x_i^s)-f_\theta(x_i^{s-})\right\|_2^2
+
m
\Big]_{+},
\\[1mm]
\mathcal{L}_{s}
&=
\mathcal{L}_{\mathrm{ce}}
+
\beta_s\mathcal{L}_{\mathrm{tri}}.
\end{aligned}
\end{equation}
where $p_\theta(y_i^s\mid x_i^s)$ denotes the predicted probability that source sample $x_i^s$ belongs to its ground-truth identity $y_i^s$. The samples $x_i^{s+}$ and $x_i^{s-}$ denote the positive and negative instances associated with anchor $x_i^s$, respectively. The parameter $m$ is the triplet margin, $[\cdot]_{+}$ denotes the hinge function and $\beta_s$ controls the contribution of metric learning to the source-domain objective.

\subsection{Target Domain Clustering}
\label{sec:target_clustering}

After source-domain training, the encoder $f_\theta$ is applied to the target samples $\{x_j^t\}_{j=1}^{N_t}$ to extract feature
representations $z_j=f_\theta(x_j^t)\in\mathbb{R}^{d}$. The extracted features are $\ell_2$-normalized as $z_j\leftarrow z_j/\lVert z_j\rVert_2$ to obtain a consistent embedding scale for subsequent distance computation. For notational simplicity, the adaptation-round index is omitted in this subsection. Since target-domain identity annotations are unavailable, a pairwise distance matrix is constructed from the normalized target features and clustering is performed to approximate the latent identity structure. We employ DBSCAN \cite{ester1996density} because it does not require the number of target identities to be specified in advance and can identify samples that do not belong to sufficiently dense regions. The resulting target pseudo labels are generated according to Eq.~\eqref{eq:pseudo_label}.

Let
$\mathcal{I}=\{j\mid\hat{y}_j^t\neq-1\}$
denote the set of samples assigned to valid clusters, where the label
$-1$ represents a DBSCAN outlier. These outliers are excluded from the
selective partition and target-domain optimization in the current
adaptation round. For the $k$-th cluster, the corresponding index set is
defined as
$\mathcal{C}_k=\{j\in\mathcal{I}\mid\hat{y}_j^t=k\}$.
The mean representation and its normalized cluster centroid are
formulated in Eq.~\eqref{eq:centroid}:
\begin{equation}
\begin{split}
\widetilde{\mu}_k
&=
\frac{1}{|\mathcal{C}_k|}
\sum_{j \in \mathcal{C}_k} z_j,
\\[1mm]
\mu_k
&=
\frac{\widetilde{\mu}_k}
{\|\widetilde{\mu}_k\|_2}.
\end{split}
\label{eq:centroid}
\end{equation}
where $\lvert\mathcal{C}_k\rvert$ denotes the number of samples assigned
to cluster $k$, $\widetilde{\mu}_k$ is the mean representation of the
cluster members, and $\mu_k$ is the corresponding $\ell_2$-normalized
centroid. Consequently, each valid target sample $x_j^t$ is associated
with an initial pseudo-label $\hat{y}_j^t$ and the centroid
$\mu_{\hat{y}_j^t}$ of its assigned cluster.

Although clustering provides an initial supervisory structure for
target-domain adaptation, its assignments may remain unreliable because
of domain discrepancy, viewpoint variation, illumination changes,
occlusion and background clutter. These factors can cause
intra-identity fragmentation and inter-identity confusion allowing
incorrect assignments to propagate during iterative learning. CGPLR
therefore evaluates each valid pseudo label using centroid compatibility,
local neighborhood consistency, competing-cluster separation and
temporal assignment stability. The resulting multi-evidence confidence
estimation is introduced in Section~\ref{sec:confidence}.

\subsection{Multi-Evidence Confidence Estimation}
\label{sec:confidence}

The pseudo labels produced by target-domain clustering provide hard identity assignments but do not indicate their reliability. A sample may be close to its assigned centroid while remaining similarly close to a competing cluster and local agreement may be unreliable when neighboring samples contain incorrect assignments. CGPLR therefore estimates sample-level reliability using complementary structural, competitive and temporal evidence. For clarity, the adaptation round index is omitted from within-round quantities unless required explicitly. Given a target sample $x_j^t$ with normalized embedding $z_j$ and current
pseudo label $\hat{y}_j^t=k$, its structural reliability is evaluated using the assigned cluster centroid and local neighborhood. The centroid compatibility, neighborhood consistency and their fused structural score are formulated in Eq.~\eqref{eq:structural_evidence}:
\begin{equation}
\label{eq:structural_evidence}
\begin{aligned}
c_j
&=
\exp\!\left(
-\frac{\left\lVert z_j-\mu_k\right\rVert_2^2}{\tau_c}
\right),
\\[1mm]
s_j
&=
\frac{1}{|\mathcal{N}_j|}
\sum_{l\in\mathcal{N}_j}
\mathbb{I}\!\left(
\hat{y}_l^t=\hat{y}_j^t
\right),
\\[1mm]
r_j
&=
\alpha c_j+(1-\alpha)s_j.
\end{aligned}
\end{equation}
where $\mathcal{N}_j\subseteq\mathcal{I}\setminus\{j\}$ denotes the set
of the $K$ nearest valid target samples to $z_j$, and
$\mathbb{I}(\cdot)$ is the indicator function. If fewer than $K$ valid
neighbors are available, all available valid neighbors are used. The
temperature $\tau_c>0$ controls the sensitivity of centroid
compatibility to the sample--centroid distance, while
$\alpha\in[0,1]$ balances centroid and neighborhood evidence. A sample
receives a high structural score when its current assignment is
consistently supported by both the assigned centroid and its local
neighborhood.

Structural reliability alone does not explicitly measure competition
from other clusters. We therefore identify the closest competing
centroid and measure its separation from the currently assigned
centroid. The strongest competing cluster, distance margin and
margin-based confidence are formulated in
Eq.~\eqref{eq:competitive_evidence}:
\begin{equation}
\label{eq:competitive_evidence}
\begin{aligned}
k_j^{-}
&=
\operatorname*{arg\,min}_{q\neq k}
\left\lVert z_j-\mu_q\right\rVert_2^2,
\\[1mm]
\Delta_j
&=
\left\lVert z_j-\mu_{k_j^{-}}\right\rVert_2^2
-
\left\lVert z_j-\mu_k\right\rVert_2^2,
\\[1mm]
m_j
&=
\sigma\!\left(
\frac{\Delta_j}{\tau_m}
\right).
\end{aligned}
\end{equation}
where $\sigma(\cdot)$ denotes the sigmoid function and $\tau_m>0$
controls the smoothness of the margin transformation. A large positive
$\Delta_j$ indicates that the assigned centroid is clearly preferred
over its strongest competitor, whereas a value close to zero indicates
an ambiguous assignment. A negative margin indicates that a competing
centroid is closer than the currently assigned centroid.

Reliability should also remain stable across successive adaptation
rounds. Let $\hat{y}_j^{t,(e)}$ denote the pseudo label of target sample
$j$ at round $e$, and let $\mu_k^{(e)}$ denote the centroid of cluster
$k$ at that round. Because cluster indices may change after
re-clustering, historical clusters are aligned with the current
clusters before their assignments are compared. Let
$\Pi_{e-h,e}$ denote the set of maximum-cardinality one-to-one mappings
between the clusters obtained at rounds $e-h$ and $e$. The
centroid-based cluster alignment and temporal stability are jointly
formulated in Eq.~\eqref{eq:temporal_stability}:
\begin{equation}
\label{eq:temporal_stability}
\begin{aligned}
\pi_{e-h\rightarrow e}
&=
\operatorname*{arg\,min}_{\pi\in\Pi_{e-h,e}}
\sum_{k\in\operatorname{dom}(\pi)}
\left\lVert
\mu_k^{(e-h)}
-
\mu_{\pi(k)}^{(e)}
\right\rVert_2^2,
\\[1mm]
t_j^{(e)}
&=
\frac{1}{H_e}
\sum_{h=1}^{H_e}
\mathbb{I}\!\left(
\pi_{e-h\rightarrow e}
\!\left(
\hat{y}_j^{t,(e-h)}
\right)
=
\hat{y}_j^{t,(e)}
\right),
\\[1mm]
H_e
&=
\min(H,e-1),
\qquad e>1.
\end{aligned}
\end{equation}
where $\operatorname{dom}(\pi)$ denotes the set of historical clusters included in the one-to-one alignment and $H$ is the maximum number of previous adaptation rounds retained in the temporal window. Historical outlier assignments and clusters that are not included in the alignment contribute zero to the corresponding indicator. A high value of $t_j^{(e)}$ indicates that the sample has maintained a stable cluster assignment over recent adaptation rounds.

Finally, the structural, competitive and temporal cues are combined into the sample-level confidence score. Since temporal evidence is not available during the first adaptation round, the structural and margin weights are renormalized in that round. The complete confidence formulation is given in Eq.~\eqref{eq:final_confidence}:
\begin{equation}
\label{eq:final_confidence}
w_j^{(e)}
=
\begin{cases}
\displaystyle
\frac{
\omega_r r_j+\omega_m m_j
}{
\omega_r+\omega_m
},
& e=1,
\\[4mm]
\omega_r r_j
+
\omega_m m_j
+
\omega_t t_j^{(e)},
& e>1,
\end{cases}
\end{equation}
where $\omega_r,\omega_m,\omega_t\geq0$, $\omega_r+\omega_m+\omega_t=1$, and $\omega_r+\omega_m>0$. The resulting score $w_j^{(e)}\in[0,1]$ serves as a bounded surrogate for the latent pseudo-label correctness indicator $\delta_j=\mathbb{I}(\hat{y}_j^t=y_j^t)$ introduced in
Section~\ref{sec:problem_formulation}. Samples consistently supported
by the structural, competitive, and temporal evidence receive higher
confidence, whereas disagreement among these cues results in more
cautious treatment. The estimated confidence scores subsequently guide
the cluster-adaptive preserve, reassign and defer decisions introduced
in the following subsection.

\subsection{Adaptive Selective Pseudo-Label Refinement}
\label{sec:label_refinement}

The pseudo labels generated through clustering may contain incorrect assignments because target-domain representations remain imperfect during adaptation. Directly relabeling every low-confidence sample can introduce additional errors, particularly when no alternative cluster is sufficiently supported. We therefore formulate pseudo-label refinement as a selective decision process in which each valid target sample is preserved, reassigned or temporarily deferred according to its confidence and the evidence supporting alternative assignments. A single global threshold may not adequately represent the heterogeneous
reliability distributions of different target clusters. For cluster
$\mathcal{C}_k^{(e)}$, its median confidence and median absolute
deviation are formulated in
Eq.~\eqref{eq:cluster_confidence_statistics}:
\begin{equation}
\label{eq:cluster_confidence_statistics}
\begin{aligned}
\bar{w}_k^{(e)}
&=
\operatorname{median}_{j\in\mathcal{C}_k^{(e)}}
w_j^{(e)},
\\[1mm]
d_k^{(e)}
&=
\operatorname{median}_{j\in\mathcal{C}_k^{(e)}}
\left|
w_j^{(e)}-\bar{w}_k^{(e)}
\right|.
\end{aligned}
\end{equation}
where $\bar{w}_k^{(e)}$ represents the central reliability level of
cluster $k$, whereas $d_k^{(e)}$ measures the dispersion of its
confidence scores.

To avoid a degenerate threshold interval when the confidence dispersion
of a cluster is zero or extremely small, a lower-bounded dispersion is
first defined as
$\widetilde{d}_k^{(e)}=\max(d_k^{(e)},\epsilon_d)$, where
$\epsilon_d>0$ is a small constant. The corresponding upper and lower
cluster-adaptive thresholds are then defined in
Eq.~\eqref{eq:adaptive_thresholds}:
\begin{equation}
\label{eq:adaptive_thresholds}
\begin{aligned}
\widetilde{d}_k^{(e)}
&=
\max\!\left(d_k^{(e)},\epsilon_d\right),
\\[1mm]
\delta_{k}^{H,(e)}
&=
\operatorname{clip}\!\left(
\bar{w}_k^{(e)}
+
\rho_H\widetilde{d}_k^{(e)},
0,1
\right),
\\[1mm]
\delta_{k}^{L,(e)}
&=
\operatorname{clip}\!\left(
\bar{w}_k^{(e)}
-
\rho_L\widetilde{d}_k^{(e)},
0,1
\right).
\end{aligned}
\end{equation}
The parameters $\rho_H$ and $\rho_L$ control the selectivity of the
upper and lower thresholds, respectively, while
$\operatorname{clip}(\cdot,0,1)$ restricts the resulting values to
$[0,1]$. These thresholds adapt the decision criteria to the reliability
distribution of each cluster rather than applying a single fixed
criterion to all target identities.

For a sample whose current assignment is not sufficiently reliable, we
evaluate whether another cluster provides stronger support. For each
candidate cluster $q$, the neighborhood support, centroid compatibility
and fused assignment evidence are formulated in
Eq.~\eqref{eq:candidate_evidence}:
\begin{equation}
\label{eq:candidate_evidence}
\begin{aligned}
a_{j,q}
&=
\frac{1}{|\mathcal{N}_j|}
\sum_{l\in\mathcal{N}_j}
\mathbb{I}\!\left(
\hat{y}_l^t=q
\right),
\\[1mm]
g_{j,q}
&=
\exp\!\left(
-\frac{
\left\lVert z_j-\mu_q\right\rVert_2^2
}{
\tau_r
}
\right),
\\[1mm]
u_{j,q}
&=
\beta g_{j,q}
+
(1-\beta)a_{j,q}.
\end{aligned}
\end{equation}
where $a_{j,q}$ measures the local support for candidate cluster $q$,
whereas $g_{j,q}$ measures the geometric compatibility of the sample
with centroid $\mu_q$. The temperature $\tau_r>0$ controls the
sensitivity of centroid compatibility to the sample--centroid distance,
and $\beta\in[0,1]$ balances geometric and neighborhood evidence. The
fused score therefore favors candidate clusters that receive consistent
support from both sources.

For a sample currently assigned to
$k=\hat{y}_j^{t,(e)}$, the best-supported alternative cluster and its
evidence gain over the current assignment are defined in
Eq.~\eqref{eq:best_candidate}:
\begin{equation}
\label{eq:best_candidate}
\begin{aligned}
q_j^{*}
&=
\operatorname*{arg\,max}_{q\neq k}
u_{j,q},
\\[1mm]
\Gamma_j
&=
u_{j,q_j^{*}}
-
u_{j,k}.
\end{aligned}
\end{equation}
A sufficiently large positive value of $\Gamma_j$ indicates that the
alternative cluster $q_j^{*}$ is more strongly supported than the
current assignment.

Using the cluster-adaptive thresholds and alternative-assignment
evidence, the preserved, reassigned and deferred subsets are formulated
in Eq.~\eqref{eq:selective_partition}:
\begin{equation}
\label{eq:selective_partition}
\begin{aligned}
\mathcal{H}^{(e)}
&=
\left\{
j\in\mathcal{I}^{(e)}
\ \middle|\
w_j^{(e)}
\geq
\delta_{\hat{y}_j^{t,(e)}}^{H,(e)}
\right\},
\\[1mm]
\mathcal{R}^{(e)}
&=
\left\{
j\in\mathcal{I}^{(e)}
\ \middle|\
\begin{gathered}
\delta_{\hat{y}_j^{t,(e)}}^{L,(e)}
\leq
w_j^{(e)}
<
\delta_{\hat{y}_j^{t,(e)}}^{H,(e)},
\\
\Gamma_j\geq\xi,
\qquad
a_{j,q_j^{*}}\geq\zeta
\end{gathered}
\right\},
\\[1mm]
\mathcal{D}^{(e)}
&=
\mathcal{I}^{(e)}
\setminus
\left(
\mathcal{H}^{(e)}
\cup
\mathcal{R}^{(e)}
\right).
\end{aligned}
\end{equation}
where $\mathcal{H}^{(e)}$, $\mathcal{R}^{(e)}$, and
$\mathcal{D}^{(e)}$ denote the preserved, reassigned and deferred
subsets, respectively. The parameter $\xi$ specifies the minimum
evidence gain required for reassignment, whereas $\zeta$ enforces
sufficient neighborhood support for the proposed alternative cluster.

Samples below the lower threshold are deferred directly because their
representations are considered insufficiently reliable for selecting a
trustworthy alternative identity. Reassignment is therefore restricted
to the intermediate-confidence region and is accepted only when both
the relative evidence gain and neighborhood-support conditions are
satisfied. Intermediate-confidence samples that fail either condition
are also deferred.

The resulting refined pseudo-label assignment is formulated in
Eq.~\eqref{eq:final_label}:
\begin{equation}
\label{eq:final_label}
\hat{y}_j^{t,*,(e)}
=
\begin{cases}
\hat{y}_j^{t,(e)},
& j\in\mathcal{H}^{(e)},
\\[2mm]
q_j^{*},
& j\in\mathcal{R}^{(e)},
\\[2mm]
\varnothing,
& j\in\mathcal{D}^{(e)}.
\end{cases}
\end{equation}
Reliable assignments are therefore preserved, recoverable assignments
are replaced by their best-supported alternatives, and ambiguous samples
receive no hard identity label during the current adaptation round.

Deferral is important because low confidence does not necessarily imply
that a particular alternative cluster is correct. Rather than forcing
ambiguous samples into potentially incorrect identities, CGPLR allows
their assignments to be reconsidered after the representation and
cluster structure are updated in subsequent adaptation rounds. Deferred
samples are excluded from hard pseudo-label classification and metric
learning but remain involved through label-free consistency
regularization, as described in
Section~\ref{sec:confidence_learning}.

The selective formulation has two defining properties. First, its
decision thresholds adapt to each cluster's reliability distribution.
Second, reassignment is permitted only when relative evidence gain and
neighborhood support jointly favor an alternative cluster; otherwise,
the sample is deferred. This design enables the correction of
recoverable assignments while limiting error propagation from ambiguous
pseudo labels.

\subsection{Reliability-Aware Selective Learning}
\label{sec:confidence_learning}

The selective refinement strategy introduced in
Section~\ref{sec:label_refinement} partitions the valid target samples
into preserved, reassigned, and deferred subsets. These subsets should
not contribute to target-domain optimization in the same manner.
Preserved samples receive reliable hard pseudo-label supervision,
reassigned samples receive more cautious corrected-label supervision,
and deferred samples are excluded from identity-based losses because no
sufficiently reliable identity assignment is available.

Let $B_t^{(e)}\subseteq\mathcal{I}^{(e)}$ denote a target mini-batch at
adaptation round $e$. Its hard-labeled and deferred portions are defined
in Eq.~\eqref{eq:learning_partitions}:
\begin{equation}
\label{eq:learning_partitions}
\begin{aligned}
B_{\mathrm{L}}^{(e)}
&=
B_t^{(e)}
\cap
\left(
\mathcal{H}^{(e)}
\cup
\mathcal{R}^{(e)}
\right),
\\[1mm]
B_{\mathrm{D}}^{(e)}
&=
B_t^{(e)}
\cap
\mathcal{D}^{(e)}.
\end{aligned}
\end{equation}
where $B_{\mathrm{L}}^{(e)}$ contains the preserved and reassigned
samples with refined hard pseudo labels, whereas
$B_{\mathrm{D}}^{(e)}$ contains the samples deferred from hard identity
supervision.

For a preserved sample, $w_j^{(e)}$ directly represents the reliability
of its retained clustering assignment. For a reassigned sample, however,
the original confidence should not be transferred directly to the new
identity. We therefore compute an alternative-specific confidence from
the relative evidence supporting the new and original assignments. The
reassignment confidence and the resulting selective learning weight are
defined in Eq.~\eqref{eq:selective_weights}:
\begin{equation}
\label{eq:selective_weights}
\begin{aligned}
\widetilde{u}_j^{(e)}
&=
\frac{
u_{j,q_j^{*}}
}{
u_{j,q_j^{*}}
+
u_{j,\hat{y}_j^{t,(e)}}
+
\epsilon
},
\qquad
j\in\mathcal{R}^{(e)},
\\[1mm]
\widetilde{w}_j^{(e)}
&=
\begin{cases}
w_j^{(e)},
& j\in\mathcal{H}^{(e)},
\\[1mm]
\eta_r\widetilde{u}_j^{(e)},
& j\in\mathcal{R}^{(e)}.
\end{cases}
\end{aligned}
\end{equation}
where $u_{j,q_j^{*}}$ and
$u_{j,\hat{y}_j^{t,(e)}}$ denote the evidence supporting the
alternative and original assignments, respectively, as formulated in
Eq.~\eqref{eq:candidate_evidence}. The constant $\epsilon>0$ ensures
numerical stability, while $\eta_r\in(0,1]$ limits the contribution of
reassigned samples so that their corrected labels are used more
conservatively.

Let $\mathcal{K}^{(e)}$ denote the set of valid pseudo identities in the
current adaptation round, and let
$\ell_{\theta,q}(x_j^t)$ denote the prediction logit associated with
target identity $q$. The corresponding target-domain identity
probability is given by
\[
p_\theta(q\mid x_j^t)
=
\frac{
\exp\!\left(\ell_{\theta,q}(x_j^t)\right)
}{
\displaystyle
\sum_{v\in\mathcal{K}^{(e)}}
\exp\!\left(\ell_{\theta,v}(x_j^t)\right)
}.
\]
Using the refined pseudo labels from Eq.~\eqref{eq:final_label}, the
reliability-weighted cross-entropy loss is formulated in
Eq.~\eqref{eq:weighted_ce}:
\begin{equation}
\label{eq:weighted_ce}
\begin{aligned}
\mathcal{L}_{\mathrm{w\text{-}ce}}
&=
-\frac{1}{Z_{\mathrm{ce}}}
\sum_{j\in B_{\mathrm{L}}^{(e)}}
\widetilde{w}_j^{(e)}
\log
p_{\theta}\!\left(
\hat{y}_j^{t,*,(e)}
\mid
x_j^t
\right),
\\[1mm]
Z_{\mathrm{ce}}
&=
\sum_{j\in B_{\mathrm{L}}^{(e)}}
\widetilde{w}_j^{(e)}
+
\epsilon.
\end{aligned}
\end{equation}
This formulation assigns greater influence to samples with more reliable
preserved or corrected labels while limiting the effect of uncertain
assignments.

To preserve the discriminative structure of the target embedding space,
we additionally employ reliability-weighted metric learning. Let
\[
d_\theta(x,x')
=
\left\|
\bar{f}_\theta(x)-\bar{f}_\theta(x')
\right\|_2^2
\]
denote the squared Euclidean distance between normalized embeddings,
where
$\bar{f}_\theta(x)=f_\theta(x)/\|f_\theta(x)\|_2$.
For each eligible anchor $j\in B_{\mathrm{L}}^{(e)}$, the positive and
negative sets are defined as
\[
\mathcal{P}_j^{(e)}
=
\left\{
l\in B_{\mathrm{L}}^{(e)}\setminus\{j\}
\ \middle|\
\hat{y}_l^{t,*,(e)}
=
\hat{y}_j^{t,*,(e)}
\right\},
\]
and
\[
\mathcal{N}_j^{(e)}
=
\left\{
l\in B_{\mathrm{L}}^{(e)}
\ \middle|\
\hat{y}_l^{t,*,(e)}
\neq
\hat{y}_j^{t,*,(e)}
\right\}.
\]
The eligible anchor set is
\[
\mathcal{A}_B^{(e)}
=
\left\{
j\in B_{\mathrm{L}}^{(e)}
\ \middle|\
\mathcal{P}_j^{(e)}\neq\varnothing,\;
\mathcal{N}_j^{(e)}\neq\varnothing
\right\}.
\]
Using batch-hard mining, the positive and negative instances are selected
as
\[
x_j^{t+}
=
\operatorname*{arg\,max}_{l\in\mathcal{P}_j^{(e)}}
d_\theta(x_j^t,x_l^t),
\qquad
x_j^{t-}
=
\operatorname*{arg\,min}_{l\in\mathcal{N}_j^{(e)}}
d_\theta(x_j^t,x_l^t).
\]
The corresponding reliability-weighted triplet objective is formulated
in Eq.~\eqref{eq:weighted_triplet}:
\begin{equation}
\label{eq:weighted_triplet}
\begin{aligned}
\mathcal{L}_{\mathrm{w\text{-}tri}}
&=
\frac{1}{Z_{\mathrm{tri}}}
\sum_{j\in\mathcal{A}_B^{(e)}}
\widetilde{w}_j^{(e)}
\left[
d_{\theta}(x_j^t,x_j^{t+})
-
d_{\theta}(x_j^t,x_j^{t-})
+
m
\right]_{+},
\\[1mm]
Z_{\mathrm{tri}}
&=
\sum_{j\in\mathcal{A}_B^{(e)}}
\widetilde{w}_j^{(e)}
+
\epsilon.
\end{aligned}
\end{equation}
where $m>0$ denotes the triplet margin and $[\cdot]_{+}$ represents the
hinge function. The reliability weight is applied to each anchor so that
samples with less reliable refined labels exert a smaller influence on
metric learning.

Deferred samples receive no hard identity label during the current
adaptation round. Rather than discarding them, CGPLR encourages
representation consistency under weak and strong image augmentations.
Let $\mathcal{A}_{w}(\cdot)$ and $\mathcal{A}_{s}(\cdot)$ denote the
weak and strong augmentation operators, respectively. Both views are
processed by the same encoder, while the weak-view representation is
treated as a fixed target through stop-gradient. The deferred-sample
consistency objective is formulated in
Eq.~\eqref{eq:deferred_consistency}:
\begin{equation}
\label{eq:deferred_consistency}
\mathcal{L}_{\mathrm{def}}
=
\frac{1}{
|B_{\mathrm{D}}^{(e)}|
+
\epsilon
}
\sum_{j\in B_{\mathrm{D}}^{(e)}}
\left\|
\operatorname{sg}\!\left[
\bar{f}_{\theta}\!\left(
\mathcal{A}_{w}(x_j^t)
\right)
\right]
-
\bar{f}_{\theta}\!\left(
\mathcal{A}_{s}(x_j^t)
\right)
\right\|_2^2.
\end{equation}
where $\operatorname{sg}[\cdot]$ denotes the stop-gradient operation.
This objective retains information from deferred samples without forcing
them into potentially incorrect identity clusters. When
$B_{\mathrm{D}}^{(e)}$ is empty, $\mathcal{L}_{\mathrm{def}}$ is set
to zero. Similarly, the corresponding classification or metric-learning
loss is set to zero when its required learning subset is empty.

The complete optimization procedure of CGPLR, including source-domain initialization, iterative target-domain clustering,
multi-evidence confidence estimation, selective pseudo-label refinement, reliability-aware target optimization and temporal memory updating is summarized in
Algorithm~\ref{alg:framework}.

\SetKwFor{ForAll}{for all}{do}{end}

\begin{algorithm}[!t]
\caption{Training Procedure of the Proposed CGPLR Framework}
\label{alg:framework}

\scriptsize
\DontPrintSemicolon
\SetAlgoVlined

\SetInd{0.50em}{0.45em}

\KwIn{%
\parbox[t]{0.78\linewidth}{%
Labeled source data $\mathcal{D}_s$ and unlabeled target data
$\mathcal{D}_t$;\\
encoder $f_\theta$ and clustering operator $\mathcal{C}(\cdot)$;\\
neighborhood size $K$, temporal window $H$, source epochs $T_s$,
and target adaptation rounds $T_t$;\\
confidence, refinement, and learning parameters
$\Theta_c$, $\Theta_r$, and $\Theta_l$.
}}

\KwOut{%
\parbox[t]{0.78\linewidth}{%
Adapted target-domain encoder $f_{\theta^*}$.
}}

\BlankLine
\textbf{Stage 1: Source-Domain Initialization}\;

Optimize $\mathcal{L}_s$ on $\mathcal{D}_s$ for $T_s$ epochs using
Eq.~\eqref{eq:source_training}, and initialize
$\mathcal{M}\leftarrow\varnothing$\;

\BlankLine

\For{$e\leftarrow1$ \KwTo $T_t$}{

\textbf{Stage 2: Target Feature Extraction and Clustering}\;

Extract and normalize the target features:
$\displaystyle
z_j^{(e)}
=
\frac{f_\theta(x_j^t)}
{\|f_\theta(x_j^t)\|_2},
\quad j=1,\ldots,N_t
$\;

Generate the pseudo labels
$\hat{\mathbf{y}}^{\,t,(e)}$ using
Eq.~\eqref{eq:pseudo_label}; form the valid sample set
$\mathcal{I}^{(e)}$ and clusters
$\mathcal{C}_k^{(e)}$\;

Compute the cluster means
$\widetilde{\mu}_k^{(e)}$ and normalized centroids
$\mu_k^{(e)}$ using Eq.~\eqref{eq:centroid}\;

\If{$e>1$}{
Set $H_e\leftarrow\min(H,e-1)$ and obtain the
minimum-cost centroid mappings
$\pi_{e-h\rightarrow e}$,
$h=1,\ldots,H_e$, using
Eq.~\eqref{eq:temporal_stability}\;
}

\BlankLine
\textbf{Stage 3: Multi-Evidence Confidence Estimation}\;

\ForAll{$j\in\mathcal{I}^{(e)}$}{

Compute the structural evidence
$(c_j^{(e)},s_j^{(e)},r_j^{(e)})$ using
Eq.~\eqref{eq:structural_evidence}\;

Compute the competing-cluster evidence
$(k_j^{-},\Delta_j^{(e)},m_j^{(e)})$ using
Eq.~\eqref{eq:competitive_evidence}\;

\If{$e>1$}{
Compute the temporal stability
$t_j^{(e)}$ using
Eq.~\eqref{eq:temporal_stability}\;
}

Compute the sample confidence
$w_j^{(e)}$ using
Eq.~\eqref{eq:final_confidence}\;
}

\ForAll{clusters $\mathcal{C}_k^{(e)}$}{

Compute the cluster-confidence statistics and the
adaptive thresholds
$\delta_k^{H,(e)}$ and $\delta_k^{L,(e)}$ using
Eqs.~\eqref{eq:cluster_confidence_statistics}
and~\eqref{eq:adaptive_thresholds}\;
}

\BlankLine
\textbf{Stage 4: Selective Pseudo-Label Refinement}\;

\ForAll{$j\in\mathcal{I}^{(e)}$}{

Evaluate the candidate clusters and obtain
$(q_j^*,\Gamma_j)$ using
Eqs.~\eqref{eq:candidate_evidence}
and~\eqref{eq:best_candidate}\;
}

Partition the valid samples into
$\mathcal{H}^{(e)}$, $\mathcal{R}^{(e)}$, and
$\mathcal{D}^{(e)}$ using
Eq.~\eqref{eq:selective_partition}\;

Assign the refined pseudo labels
$\hat{y}_j^{t,*,(e)}$ using
Eq.~\eqref{eq:final_label}\;

\BlankLine
\textbf{Stage 5: Confidence-Aware Target Optimization}\;

\ForAll{target mini-batches $B_t$}{

Construct the learning and deferred subsets, and compute
the selective weights
$\widetilde{w}_j^{(e)}$ using
Eqs.~\eqref{eq:learning_partitions}
and~\eqref{eq:selective_weights}\;

Compute
$\mathcal{L}_{\mathrm{w\text{-}ce}}$,
$\mathcal{L}_{\mathrm{w\text{-}tri}}$, and
$\mathcal{L}_{\mathrm{def}}$ using
Eqs.~\eqref{eq:weighted_ce}--%
\eqref{eq:deferred_consistency}\;

Form $\mathcal{L}_t$ using
Eq.~\eqref{eq:target_loss}, and update
$\theta\leftarrow
\theta-\eta\nabla_\theta\mathcal{L}_t$\;
}

\BlankLine
\textbf{Stage 6: Temporal Memory Update}\;

Update the temporal memory:
$\displaystyle
\mathcal{M}
\leftarrow
\operatorname{Tail}_{H}
\left(
\mathcal{M},
\hat{\mathbf{y}}^{\,t,(e)},
\{\mu_k^{(e)}\}
\right)
$\;

}

Set $\theta^*\leftarrow\theta$\;

\Return{$f_{\theta^*}$}\;

\end{algorithm}

Finally, the complete reliability-aware target-domain objective is
formulated in Eq.~\eqref{eq:target_loss}:
\begin{equation}
\label{eq:target_loss}
\mathcal{L}_{t}
=
\mathcal{L}_{\mathrm{w\text{-}ce}}
+
\lambda_{\mathrm{tri}}
\mathcal{L}_{\mathrm{w\text{-}tri}}
+
\lambda_{\mathrm{def}}
\mathcal{L}_{\mathrm{def}}.
\end{equation}
where $\lambda_{\mathrm{tri}}\geq0$ and
$\lambda_{\mathrm{def}}\geq0$ control the contributions of reliability-weighted metric learning and deferred-sample consistency, respectively. Reliability therefore determines both the strength and the type of supervision assigned to each target sample: preserved samples receive confidence-weighted hard supervision, reassigned samples receive cautious corrected-label supervision and deferred samples contribute through label-free representation consistency.

\FloatBarrier

\section{Experiments}
\label{sec:exp}

\subsection{Datasets}
\label{sec:datasets}

We evaluate CGPLR on three widely used person re-identification
benchmarks: Market-1501~\cite{zheng2015market},
MSMT17~\cite{wei2018person}, and
CUHK03-NP-D~\cite{li2014deepreid,zhong2017reranking}.
Market-1501 contains 32,668 images of 1,501 identities captured by
six cameras with 751 identities used for training and 750 for
evaluation. MSMT17 contains 126,441 images of 4,101 identities
collected by 15 cameras under substantial illumination, viewpoint,
background, and scene variations. CUHK03-NP-D contains 14,097
detected images of 1,467 identities captured by two cameras under
the new evaluation protocol, which introduces additional detection
and localization noise. The main comparisons consider
Market-1501$\rightarrow$MSMT17 and
MSMT17$\rightarrow$Market-1501, while
Market-1501$\rightarrow$CUHK03-NP-D is included as an additional
cross-domain evaluation. In all settings, identity annotations are
available only in the source domain and remain inaccessible in the
target domain during adaptation.

\subsection{Experimental Protocol}
\label{sec:experimental_protocol}

For each transfer setting, the encoder is first trained on the labeled source domain using the classification and triplet objectives in Eq.~\eqref{eq:source_training}. The source-initialized model is then adapted to the unlabeled target domain. At each adaptation round, target features are extracted and normalized, DBSCAN is applied to generate pseudo labels and the resulting assignments are processed through multi-evidence reliability estimation, cluster-adaptive selective refinement and reliability-aware selective learning. Historical pseudo labels and cluster centroids are retained only for temporal-stability estimation and are aligned with the current clustering before comparison.

Target-domain identity annotations are not used to generate pseudo labels, estimate confidence, determine adaptive thresholds, select training
samples, choose hyperparameters or update the network. They are accessed only after adaptation to report retrieval performance and to conduct
diagnostic analyses of pseudo-label correctness. Final source-to-target evaluation follows the standard single-query protocol of each benchmark.

\subsection{Evaluation Metrics}
\label{sec:evaluation_metrics}

Retrieval performance is evaluated using the cumulative matching characteristic (CMC) and mean average precision (mAP), with Rank-1 accuracy (R1) reported as the primary CMC measure and mAP assessing the quality of the complete ranked retrieval list across query identities. For diagnostic analysis, target-domain annotations are accessed only after adaptation to compute preserved-label precision, reassignment precision, correction rate, hard-label coverage, deferral rate and the concentration of pseudo-label errors in the deferred subset; they are never used for pseudo-label generation, sample selection or model optimization.

\subsection{Implementation Details}
\label{sec:implementation_details}

CGPLR is implemented in PyTorch using an ImageNet-pretrained ResNet-50 backbone. Input images are resized to $256\times128$. Random horizontal flipping and random cropping are applied during source and target training, while the deferred-sample branch additionally uses weak and strong augmentations of the same target image. The model is optimized with Adam using an initial learning rate of $3.5\times10^{-4}$, a mini-batch size of 64, and a triplet margin of
$m=0.3$. Target pseudo labels are generated using DBSCAN, and the $K=5$ nearest valid target samples are used to estimate local neighborhood consistency.
Target adaptation is conducted for $T_t=50$ rounds with feature extraction, clustering, reliability estimation, selective refinement and model optimization repeated at each round. The optimization and method-specific settings are summarized in Table~\ref{tab:hyperparameters}.

\begin{table}[!t]
\centering
\caption{Optimization and method-specific settings used by CGPLR.}
\label{tab:hyperparameters}

\scriptsize
\setlength{\tabcolsep}{2pt}
\renewcommand{\arraystretch}{0.92}

\begin{tabularx}{\columnwidth}{%
@{}
>{\raggedright\arraybackslash}p{0.34\columnwidth}
>{\centering\arraybackslash}p{0.19\columnwidth}
>{\raggedright\arraybackslash}X
@{}
}
\toprule
Parameter & Value & Description \\
\midrule

Batch size $B$
& 64
& Mini-batch size \\

Learning rate $\eta$
& $3.5\times10^{-4}$
& Initial Adam learning rate \\

Adaptation rounds $T_t$
& 50
& Target adaptation rounds \\

Triplet margin $m$
& 0.3
& Metric-learning margin \\

Neighborhood size $K$
& 5
& Number of nearest neighbors \\

Structural balance $\alpha$
& 0.5
& Balance between $c_j$ and $s_j$ \\

Temporal window $H$
& 3
& Number of retained historical rounds \\

Fusion weights $(\omega_r,\omega_m,\omega_t)$
& $(0.5,0.3,0.2)$
& Structural, margin, and temporal weights \\

Temperatures $(\tau_c,\tau_m,\tau_r)$
& $(0.5,0.1,0.5)$
& Confidence and reassignment temperatures \\

Threshold factors $(\rho_H,\rho_L)$
& $(1.0,1.0)$
& Cluster-adaptive threshold scales \\

Reassignment parameters $(\beta,\xi,\zeta)$
& $(0.5,0.1,0.6)$
& Evidence balance and acceptance criteria \\

Cautious factor $\eta_r$
& 0.7
& Weight applied to reassigned samples \\

Source triplet weight $\beta_s$
& 1.0
& Source metric-learning weight \\

Loss weights $(\lambda_{\mathrm{tri}},\lambda_{\mathrm{def}})$
& $(1.0,0.1)$
& Metric and consistency-loss weights \\

\bottomrule
\end{tabularx}
\end{table}

\subsection{Comparison with State-of-the-Art}

Table~\ref{tab:sota} compares CGPLR with recent domain-adaptive person Re-ID methods under the two main transfer settings. The results
of competing methods are taken from the corresponding publications under their reported evaluation protocols, while methods following a
different adaptation setting are identified separately. On Market-1501$\rightarrow$MSMT17, CGPLR achieves 76.4\% R1 and 47.6\% mAP, outperforming ADDNet, the strongest competing method in the table by 1.2 and 0.9 percentage points, respectively. This larger gain indicates that the selective preserve, reassign and defer strategy is particularly effective under substantial appearance and camera variations.

On MSMT17$\rightarrow$Market-1501, CGPLR obtains 95.1\% R1 and 87.7\% mAP, exceeding ADDNet by 0.8 and 0.7 points, respectively. Although the gains are smaller on the comparatively less diverse Market-1501 target domain, CGPLR consistently achieves the best performance in both directions. These results demonstrate that reliability is effective not only as a sample weight but also as a decision variable for determining whether a pseudo label should be preserved, corrected or deferred.

\begin{table}[!t]
\centering

\caption{Comparison with recent state-of-the-art domain-adaptive
person Re-ID methods. Rank-1 accuracy (R1) and mAP are reported (\%).
Best and second-best results are shown in bold and underlined, respectively.}
\label{tab:sota}

\fontsize{6.9}{7.9}\selectfont
\setlength{\tabcolsep}{4pt}
\renewcommand{\arraystretch}{0.90}

\begin{threeparttable}

\begin{adjustbox}{max width=\columnwidth}
\begin{tabular}{@{}llcccc@{}}
\toprule
\multirow{2}{*}{Venue/Year}
& \multirow{2}{*}{Method}
& \multicolumn{2}{c}{M$\rightarrow$MSMT}
& \multicolumn{2}{c}{MSMT$\rightarrow$M} \\
\cmidrule(lr){3-4}
\cmidrule(lr){5-6}
& & R1 & mAP & R1 & mAP \\
\midrule

NeurIPS'20
& SpCL~\cite{ge2020self}
& 53.7 & 26.8 & 90.3 & 76.7 \\

ICCV'21
& OPLG-HCD~\cite{zheng2021online}
& 54.9 & 28.4 & 91.4 & 80.2 \\

ICCV'21
& IDM~\cite{dai2021idm}
& 61.3 & 33.5 & 68.3 & 40.9 \\

AAAI'22
& SECRET~\cite{he2022secret}
& 60.4 & 31.3 & 92.6 & 81.0 \\

AAAI'22
& IDM+DARC~\cite{hu2022divide}
& 64.5 & 35.2 & 94.1 & 85.1 \\

AAAI'22
& MCRN~\cite{wu2022multi}
& 67.5 & 35.7 & 93.8 & 83.8 \\

ICCV'23
& CaCL~\cite{lee2023camera}
& 66.6 & 36.5 & 93.8 & 84.7 \\

JVCIR'23
& MTNet~\cite{chen2023mtnet}
& 59.6 & 32.1 & 93.0 & 82.7 \\

KBS'24
& ADDNet~\cite{wang2024attention}
& \underline{75.2}
& \underline{46.7}
& \underline{94.3}
& \underline{87.0} \\

PR'24
& FUReID~\cite{peng2024adapt}
& 52.6 & 26.2 & 92.7 & 81.7 \\

TMM'25
& SecureDA~\cite{qu2025secureda}\tnote{$\dagger$}
& 68.3 & 36.7 & 69.5 & 39.8 \\

PR'26
& ADA~\cite{zhang2025ada}
& 46.0 & 23.1 & 92.8 & 83.4 \\

\midrule

\textbf{Our Framework}
& \textbf{CGPLR}
& \textbf{76.4}
& \textbf{47.6}
& \textbf{95.1}
& \textbf{87.7} \\

\bottomrule
\end{tabular}
\end{adjustbox}

\begin{tablenotes}[flushleft]
\fontsize{6.3}{7.0}\selectfont
\item[$\dagger$] Source-free domain adaptation method, where only a
pretrained source model is used during adaptation.
\end{tablenotes}

\end{threeparttable}
\end{table}

To examine transfer beyond the two main benchmarks, we additionally evaluate CGPLR on Market-1501$\rightarrow$CUHK03-NP-D. Because this setting is used as a controlled additional evaluation rather than a direct state-of-the-art comparison, Table~\ref{tab:cuhk_additional} compares the source-only model, the clustering baseline and the complete CGPLR framework under the same experimental setting. This evaluation examines whether the proposed reliability-aware design remains effective when the target domain also contains detection-induced localization noise.

\begin{table}[!t]
\centering

\caption{Controlled additional UDA evaluation on
Market-1501$\rightarrow$CUHK03-NP-D. Rank-$k$ accuracy and mAP
are reported (\%). $\Delta$R1 and $\Delta$mAP denote performance
differences relative to the clustering baseline.}
\label{tab:cuhk_additional}

\scriptsize
\setlength{\tabcolsep}{2pt}
\renewcommand{\arraystretch}{0.80}

\resizebox{0.94\columnwidth}{!}{%
\begin{tabular}{@{}lcccccc@{}}
\toprule
Method
& R1
& R5
& R10
& mAP
& $\Delta$R1
& $\Delta$mAP \\
\midrule

Source-only baseline
& 37.2
& 54.9
& 62.7
& 23.1
& $-23.7$
& $-12.5$ \\

Clustering baseline
& 60.9
& 77.8
& 83.6
& 35.6
& $0.0$
& $0.0$ \\

\midrule

\textbf{Full CGPLR}
& \textbf{67.3}
& \textbf{83.4}
& \textbf{88.8}
& \textbf{40.7}
& \textbf{$+6.4$}
& \textbf{$+5.1$} \\

\bottomrule
\end{tabular}%
}

\end{table}

Table~\ref{tab:cuhk_additional} evaluates CGPLR on Market-1501$\rightarrow$CUHK03-NP-D using the source-only model, the clustering baseline and the complete framework. Clustering-based adaptation improves R1 from 37.2\% to 60.9\% and mAP from 23.1\% to 35.6\%. The full CGPLR framework further reaches 67.3\% R1, 83.4\% R5, 88.8\% R10 and 40.7\% mAP. These results correspond to improvements of 6.4 R1 points and 5.1 mAP points over the clustering baseline, showing that the proposed reliability-aware design remains effective under the additional localization noise introduced by detected pedestrian boxes.

\subsection{Ablation Studies}
\label{sec:ablation}

\subsubsection{Stage-wise contribution}

The stage-wise ablation in Table~\ref{tab:ablation_detailed} evaluates the progressive effect of clustering-based pseudo labeling, centroid confidence, neighborhood consistency, label refinement and confidence-aware learning. These results show that each successive stage improves the source-only and clustering baselines across all three target domains. The largest gains are obtained when clustering is introduced, while the subsequent confidence, refinement and learning stages provide complementary improvements by reducing the influence of unreliable assignments.

\begin{table}[!t]
\centering
\caption{Stage-wise ablation of CGPLR. PL, CC, NC, SR, and
RSL denote pseudo labeling, centroid compatibility, neighborhood
consistency, selective refinement, and reliability-aware selective
learning, respectively. R1 and mAP are reported (\%).}
\label{tab:ablation_detailed}

\scriptsize
\setlength{\tabcolsep}{2.2pt}
\renewcommand{\arraystretch}{0.92}

\begin{tabularx}{\columnwidth}
{@{}>{\raggedright\arraybackslash}Xccccc cc@{}}
\toprule
\multicolumn{8}{c}{\textbf{Market-1501$\rightarrow$MSMT17}} \\
\midrule
Setting & PL & CC & NC & SR & RSL & R1 & mAP \\
\midrule

Source-only baseline
& -- & -- & -- & -- & --
& 38.6 & 15.4 \\

Clustering baseline
& \checkmark & -- & -- & -- & --
& 64.8 & 35.2 \\

+ Centroid compatibility
& \checkmark & \checkmark & -- & -- & --
& 68.7 & 38.4 \\

+ Neighborhood consistency
& \checkmark & \checkmark & \checkmark & -- & --
& 71.3 & 41.2 \\

+ Selective refinement
& \checkmark & \checkmark & \checkmark & \checkmark & --
& 74.1 & 44.6 \\

\textbf{Full CGPLR}
& \checkmark
& \checkmark
& \checkmark
& \checkmark
& \checkmark
& \textbf{76.4}
& \textbf{47.6} \\

\bottomrule
\end{tabularx}

\vspace{4pt}

\begin{tabularx}{\columnwidth}
{@{}>{\raggedright\arraybackslash}Xccccc cc@{}}
\toprule
\multicolumn{8}{c}{\textbf{MSMT17$\rightarrow$Market-1501}} \\
\midrule
Setting & PL & CC & NC & SR & RSL & R1 & mAP \\
\midrule

Source-only baseline
& -- & -- & -- & -- & --
& 72.8 & 48.9 \\

Clustering baseline
& \checkmark & -- & -- & -- & --
& 89.6 & 76.4 \\

+ Centroid compatibility
& \checkmark & \checkmark & -- & -- & --
& 91.2 & 79.1 \\

+ Neighborhood consistency
& \checkmark & \checkmark & \checkmark & -- & --
& 92.7 & 82.0 \\

+ Selective refinement
& \checkmark & \checkmark & \checkmark & \checkmark & --
& 94.0 & 85.1 \\

\textbf{Full CGPLR}
& \checkmark
& \checkmark
& \checkmark
& \checkmark
& \checkmark
& \textbf{95.1}
& \textbf{87.7} \\

\bottomrule
\end{tabularx}

\vspace{4pt}

\begin{tabularx}{\columnwidth}
{@{}>{\raggedright\arraybackslash}Xccccc cc@{}}
\toprule
\multicolumn{8}{c}{\textbf{Market-1501$\rightarrow$CUHK03-NP-D}} \\
\midrule
Setting & PL & CC & NC & SR & RSL & R1 & mAP \\
\midrule

Source-only baseline
& -- & -- & -- & -- & --
& 37.2 & 23.1 \\

Clustering baseline
& \checkmark & -- & -- & -- & --
& 60.9 & 35.6 \\

+ Centroid compatibility
& \checkmark & \checkmark & -- & -- & --
& 62.8 & 37.0 \\

+ Neighborhood consistency
& \checkmark & \checkmark & \checkmark & -- & --
& 64.2 & 38.2 \\

+ Selective refinement
& \checkmark & \checkmark & \checkmark & \checkmark & --
& 65.8 & 39.4 \\

\textbf{Full CGPLR}
& \checkmark
& \checkmark
& \checkmark
& \checkmark
& \checkmark
& \textbf{67.3}
& \textbf{40.7} \\

\bottomrule
\end{tabularx}
\end{table}
\subsubsection{Mechanism-level verification}

To isolate the contribution of the individual mechanisms introduced in CGPLR, we construct controlled variants in which one mechanism is removed or replaced at a time while all remaining components and training settings are kept unchanged. This one-factor-at-a-time design allows the effect of each mechanism to be evaluated without introducing confounding changes in the remaining adaptation pipeline. Specifically, we examine the effects of competing-centroid evidence, temporal stability, cluster-adaptive thresholds, selective deferral, retaining deferred samples, deferred-sample consistency learning and cautious weighting of reassigned samples. Reporting the results under both transfer directions also shows whether the contribution of each mechanism remains consistent across different target-domain conditions.

\begin{table}[!t]
\centering

\caption{Direct ablation of the individual mechanisms
introduced in CGPLR. Rank-1 accuracy (R1) and mAP are reported (\%).
$\Delta$mAP denotes the difference relative to the full CGPLR model
under the corresponding transfer setting. Therefore, the full model
has $\Delta$mAP $=0.0$.}
\label{tab:ablation_mechanisms}

\scriptsize
\setlength{\tabcolsep}{3pt}
\renewcommand{\arraystretch}{0.95}

\begin{tabularx}{\columnwidth}
{@{}>{\raggedright\arraybackslash}Xccc@{}}
\toprule
\multicolumn{4}{c}{\textbf{Market-1501$\rightarrow$MSMT17}} \\
\midrule
Variant
& R1
& mAP
& \shortstack{$\Delta$mAP\\vs. full} \\
\midrule

Without competing-centroid margin
& 74.9
& 45.7
& $-1.9$ \\

Without temporal stability
& 75.3
& 46.2
& $-1.4$ \\

Global thresholds instead of cluster-adaptive thresholds
& 74.6
& 45.2
& $-2.4$ \\

Forced reassignment without deferral
& 73.8
& 44.1
& $-3.5$ \\

Discard deferred samples
& 75.0
& 45.8
& $-1.8$ \\

Without deferred consistency loss
& 75.5
& 46.5
& $-1.1$ \\

Equal weights for preserved and reassigned samples
& 75.1
& 46.0
& $-1.6$ \\

\midrule

\textbf{Full CGPLR}
& \textbf{76.4}
& \textbf{47.6}
& \textbf{0.0} \\

\bottomrule
\end{tabularx}

\vspace{5pt}

\begin{tabularx}{\columnwidth}
{@{}>{\raggedright\arraybackslash}Xccc@{}}
\toprule
\multicolumn{4}{c}{\textbf{MSMT17$\rightarrow$Market-1501}} \\
\midrule
Variant
& R1
& mAP
& \shortstack{$\Delta$mAP\\vs. full} \\
\midrule

Without competing-centroid margin
& 94.4
& 86.5
& $-1.2$ \\

Without temporal stability
& 94.6
& 86.9
& $-0.8$ \\

Global thresholds instead of cluster-adaptive thresholds
& 94.2
& 86.1
& $-1.6$ \\

Forced reassignment without deferral
& 93.8
& 85.4
& $-2.3$ \\

Discard deferred samples
& 94.3
& 86.6
& $-1.1$ \\

Without deferred consistency loss
& 94.7
& 87.1
& $-0.6$ \\

Equal weights for preserved and reassigned samples
& 94.5
& 86.8
& $-0.9$ \\

\midrule

\textbf{Full CGPLR}
& \textbf{95.1}
& \textbf{87.7}
& \textbf{0.0} \\

\bottomrule
\end{tabularx}

\end{table}

Table~\ref{tab:ablation_mechanisms} directly evaluates the mechanisms introduced in CGPLR. Removing the competing-centroid margin decreases mAP by 1.9 and 1.2 points on Market-1501$\rightarrow$MSMT17 and MSMT17$\rightarrow$Market-1501, respectively, confirming that assigned-centroid compatibility alone cannot reliably identify samples that are also close to competing clusters. Excluding temporal stability causes consistent reductions of 1.4 and 0.8 points, showing that assignment history provides complementary evidence for distinguishing stable and fluctuating pseudo labels. Replacing cluster-adaptive thresholds with global thresholds reduces mAP by 2.4 and 1.6 points, indicating that a uniform confidence criterion cannot adequately handle clusters with different reliability distributions.

The largest degradation occurs when uncertain samples are forcibly reassigned without deferral, producing mAP drops of 3.5 and 2.3 points. This result confirms that ambiguous samples should not receive a hard alternative identity unless sufficient evidence is available. Completely discarding deferred samples also reduces performance by 1.8 and 1.1 points, whereas removing only the deferred consistency loss causes smaller drops of 1.1 and 0.6 points. Together, these results show that deferred samples should not be discarded and that label-free consistency learning provides an additional benefit while they remain excluded from identity-based supervision. Finally, equal weighting of preserved and reassigned samples reduces mAP by 1.6 and 0.9 points, supporting more cautious supervision for corrected labels. Overall, the full CGPLR model performs best in both transfer settings demonstrating that the proposed mechanisms are complementary.

\subsection{Pseudo-Label Reliability and Selective Decision Analysis}

To evaluate the selective mechanism beyond retrieval accuracy, we examine the preserved, reassigned and deferred subsets in Table~\ref{tab:selective_quality}. Let $\mathcal{I}^{(e)}$ denote the valid clustered target samples at adaptation round $e$, and let $\mathcal{H}^{(e)}$, $\mathcal{R}^{(e)}$, and $\mathcal{D}^{(e)}$ denote the preserved, reassigned and deferred subsets, respectively. Target identity labels are used only for post-hoc diagnosis and are never accessed during adaptation. The decision-level metrics are defined as

\allowdisplaybreaks[4]
\begin{align}
\mathrm{Prec}_{\mathrm{pre}}
&=
\frac{
\sum_{j\in\mathcal{H}^{(e)}}
\mathbb{I}(\hat{y}^{t,*}_{j}=y_j^t)
}{
|\mathcal{H}^{(e)}|
}\times100,
\label{eq:preserved_precision}
\\
\mathrm{Prec}_{\mathrm{rea}}
&=
\frac{
\sum_{j\in\mathcal{R}^{(e)}}
\mathbb{I}(\hat{y}^{t,*}_{j}=y_j^t)
}{
|\mathcal{R}^{(e)}|
}\times100,
\label{eq:reassignment_precision}
\\
\mathrm{Corr}
&=
\frac{
\sum_{j\in\mathcal{R}^{(e)}}
\mathbb{I}(\hat{y}^{t}_{j}\neq y_j^t)
\mathbb{I}(\hat{y}^{t,*}_{j}=y_j^t)
}{
\sum_{j\in\mathcal{R}^{(e)}}
\mathbb{I}(\hat{y}^{t}_{j}\neq y_j^t)
}\times100,
\label{eq:correction_rate}
\\
\mathrm{Coverage}_{\mathrm{hard}}
&=
\frac{
|\mathcal{H}^{(e)}|+|\mathcal{R}^{(e)}|
}{
|\mathcal{I}^{(e)}|
}\times100,
\label{eq:hard_label_coverage}
\\
\mathrm{Rate}_{\mathrm{def}}
&=
\frac{
|\mathcal{D}^{(e)}|
}{
|\mathcal{I}^{(e)}|
}\times100,
\label{eq:deferral_rate}
\\
\mathrm{Err}_{\mathrm{def}}
&=
\frac{
\sum_{j\in\mathcal{D}^{(e)}}
\mathbb{I}(\hat{y}^{t}_{j}\neq y_j^t)
}{
|\mathcal{D}^{(e)}|
}\times100.
\label{eq:deferred_error_rate}
\end{align}
\allowdisplaybreaks[0]

\begin{table}[!t]
\centering
\caption{Decision-level pseudo-label analysis on MSMT17$\rightarrow$Market-1501. Target labels are used only for post-hoc diagnosis; statistics are averaged over the final five adaptation rounds (\%).}
\label{tab:selective_quality}

\scriptsize
\setlength{\tabcolsep}{1.8pt}
\renewcommand{\arraystretch}{0.94}

\begin{tabularx}{\columnwidth}
{@{}>{\raggedright\arraybackslash}X
>{\centering\arraybackslash}p{0.100\columnwidth}
>{\centering\arraybackslash}p{0.145\columnwidth}
>{\centering\arraybackslash}p{0.155\columnwidth}
>{\centering\arraybackslash}p{0.125\columnwidth}@{}}
\toprule

Metric
& \shortstack{Clustering\\baseline}
& \shortstack{Multi-evidence\\reliability}
& \shortstack{+ Cluster-adaptive\\refinement}
& \shortstack{\textbf{Full}\\\textbf{CGPLR}} \\

\midrule

Preserved precision
& 89.4 & 96.2 & 97.0 & \textbf{97.8} \\

Reassignment precision
& -- & -- & 84.7 & \textbf{87.3} \\

Correction rate
& -- & -- & 67.5 & \textbf{72.6} \\

Preserved rate
& 100.0 & 81.4 & 78.2 & \textbf{79.1} \\

Reassignment rate
& 0.0 & 0.0 & 9.8 & \textbf{10.5} \\

Hard-label coverage
& 100.0 & 81.4 & 88.0 & \textbf{89.6} \\

Deferral rate
& 0.0 & 18.6 & 12.0 & \textbf{10.4} \\

Deferred error rate
& -- & 43.8 & 65.3 & \textbf{70.8} \\

\bottomrule
\end{tabularx}

\vspace{0.5mm}

\begin{minipage}{0.99\columnwidth}
\scriptsize
\textit{Note:} A dash denotes an undefined conditional metric because the required reassigned or deferred subset is empty; 0.0 indicates that no samples belong to that subset.
\end{minipage}

\end{table}

Table~\ref{tab:selective_quality} confirms that the selective mechanism separates reliable, correctable and highly ambiguous assignments as intended. Multi-evidence reliability raises preserved-label precision from 89.4\% to 96.2\% by deferring 18.6\% of less reliable samples, whose error rate is 43.8\%. Cluster-adaptive refinement recovers uncertain samples with 84.7\% reassignment precision and a 67.5\% correction rate, increasing hard-label coverage to 88.0\%. Full CGPLR provides the best balance, achieving 97.8\% preserved-label precision, 87.3\% reassignment precision, a 72.6\% correction rate and 89.6\% hard-label coverage, while the remaining 10.4\% deferred samples exhibit a 70.8\% error rate. These results show that deferral concentrates highly unreliable assignments while preserving effective pseudo-label utilization.

\subsection{Hyperparameter Sensitivity and Training Dynamics}

We evaluate the sensitivity of CGPLR to the key parameters governing temporal reliability, cluster-adaptive decision thresholds, selective
reassignment and deferred-sample learning. Each parameter is varied individually while all remaining settings are fixed to the default
configuration reported in Table~\ref{tab:hyperparameters}. The upper and lower threshold factors are varied jointly to evaluate the sensitivity to the
overall width of the cluster-adaptive decision interval. The results on MSMT17$\rightarrow$Market-1501 are presented in Table~\ref{tab:sensitivity}.

\begin{table}[!t]
\centering

\caption{Sensitivity of the revised CGPLR parameters on
MSMT17$\rightarrow$Market-1501. One parameter is varied at a time,
while all remaining parameters are fixed at their default values.
$\Delta$mAP denotes the difference from the selected configuration.}
\label{tab:sensitivity}

\fontsize{6.9}{7.5}\selectfont
\setlength{\tabcolsep}{2pt}
\renewcommand{\arraystretch}{0.90}

\begin{tabular}{@{}llccc@{}}
\toprule
Parameter
& Tested value
& R1
& mAP
& $\Delta$mAP \\
\midrule

\multirow{4}{*}{Temporal window $H$}
& 1
& 94.4
& 86.8
& $-0.9$ \\

& 2
& 94.8
& 87.3
& $-0.4$ \\

& \textbf{3}
& \textbf{95.1}
& \textbf{87.7}
& \textbf{0.0} \\

& 4
& 94.9
& 87.4
& $-0.3$ \\

\midrule

\multirow{4}{*}{%
\shortstack[l]{Threshold factors\\$(\rho_H,\rho_L)$}}
& $(0.8,0.8)$
& 94.6
& 87.0
& $-0.7$ \\

& $(0.9,0.9)$
& 94.9
& 87.4
& $-0.3$ \\

& $\mathbf{(1.0,1.0)}$
& \textbf{95.1}
& \textbf{87.7}
& \textbf{0.0} \\

& $(1.2,1.2)$
& 94.7
& 87.1
& $-0.6$ \\

\midrule

\multirow{4}{*}{Reassignment gain $\xi$}
& 0.05
& 94.7
& 87.1
& $-0.6$ \\

& \textbf{0.10}
& \textbf{95.1}
& \textbf{87.7}
& \textbf{0.0} \\

& 0.15
& 94.9
& 87.4
& $-0.3$ \\

& 0.20
& 94.5
& 86.8
& $-0.9$ \\

\midrule

\multirow{4}{*}{Temporal weight $\omega_t$}
& 0.0
& 94.2
& 86.4
& $-1.3$ \\

& 0.1
& 94.7
& 87.1
& $-0.6$ \\

& \textbf{0.2}
& \textbf{95.1}
& \textbf{87.7}
& \textbf{0.0} \\

& 0.3
& 94.8
& 87.2
& $-0.5$ \\

\midrule

\multirow{4}{*}{%
\shortstack[l]{Deferred loss\\$\lambda_{\mathrm{def}}$}}
& 0.00
& 94.4
& 86.6
& $-1.1$ \\

& 0.05
& 94.8
& 87.2
& $-0.5$ \\

& \textbf{0.10}
& \textbf{95.1}
& \textbf{87.7}
& \textbf{0.0} \\

& 0.20
& 94.6
& 86.9
& $-0.8$ \\

\bottomrule
\end{tabular}

\vspace{1mm}

\begin{minipage}{\columnwidth}
\fontsize{6.1}{6.8}\selectfont
\raggedright
\textit{Note:} When $\omega_t$ is varied, the structural and
competing-margin weights are proportionally normalized such that
$\omega_r+\omega_m+\omega_t=1$.
\end{minipage}

\end{table}

Table~\ref{tab:sensitivity} shows that CGPLR remains stable under moderate parameter variations. A temporal window of $H=3$ provides the
best balance between insufficient historical evidence and outdated assignments. The threshold factors perform best at $(\rho_H,\rho_L)=(1.0,1.0)$, while smaller or larger values reduce the separation quality among preserve, reassign and defer decisions. For reassignment, $\xi=0.1$ achieves the highest performance; a smaller value admits weak alternative assignments, whereas a larger value rejects potentially recoverable samples. Removing temporal evidence by setting $\omega_t=0$ decreases mAP by 1.3 points, confirming that assignment history complements the current structural and competing-cluster evidence. Similarly, setting $\lambda_{\mathrm{def}}=0$ reduces mAP by 1.1 points, whereas excessive consistency regularization also lowers performance. Overall, the selected configuration achieves the best result without exhibiting sharp performance degradation around the optimum.

\subsection{Qualitative and Computational Analysis}

\begin{figure*}[!t]
\centering
\includegraphics[width=0.80\textwidth]{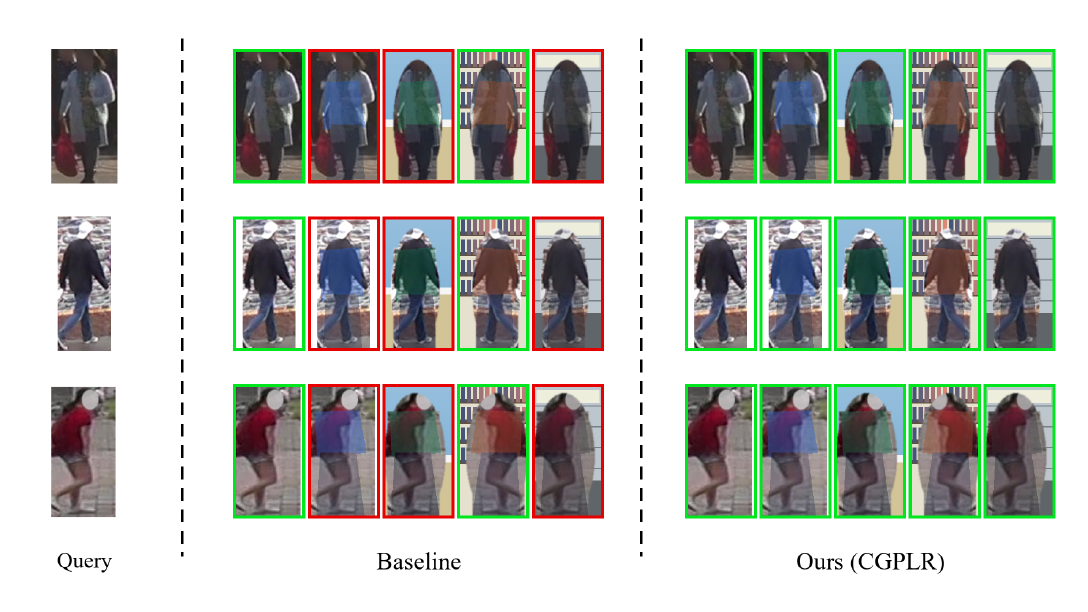}
\caption{Qualitative retrieval comparison between the clustering baseline and CGPLR under challenging target-domain appearance variations. Green and red boxes denote correct and incorrect matches, respectively.}
\label{fig:qualitative_retrieval_cgplr}
\end{figure*}

Fig.~\ref{fig:qualitative_retrieval_cgplr} compares the top-ranked retrieval results of the clustering baseline and CGPLR under illumination changes, pose variation, background clutter, changes in clothing appearance and imperfect cropping. The baseline often ranks visually similar but incorrect identities among the leading results, whereas CGPLR retrieves more correct matches and places fewer confusing identities at high ranks. The improvement is particularly evident when query and gallery images differ substantially in viewpoint or appearance. These observations indicate that the proposed refinement strategy produces more reliable retrieval rankings under challenging target-domain variations consistent with the quantitative improvements reported in the preceding experiments.

\begin{figure}[!t]
\centering
\includegraphics[width=0.98\columnwidth]{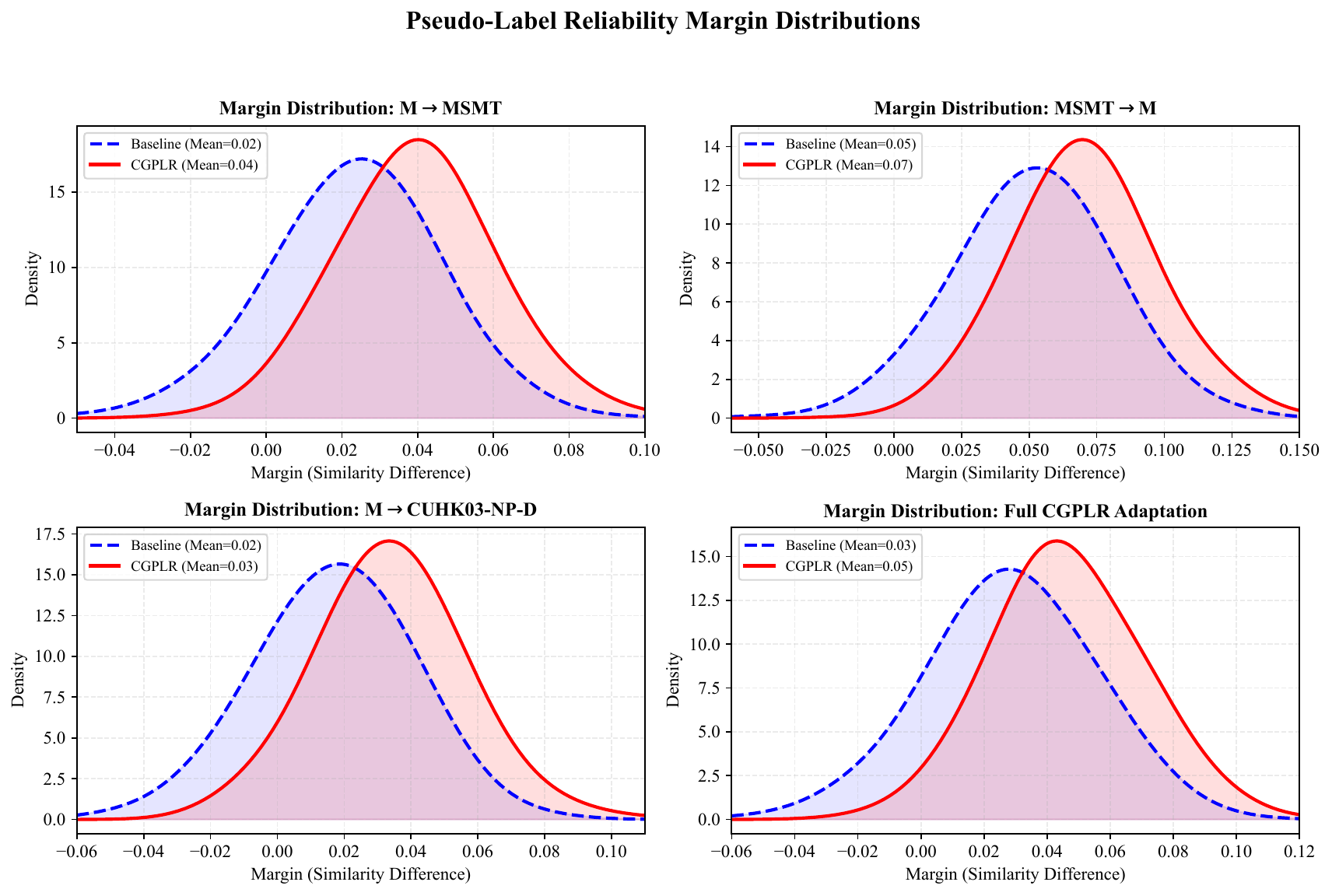}
\caption{Distributions of the competing-centroid distance margin under different cross-domain adaptation settings.}
\label{fig:margin_distribution}
\end{figure}

We further analyze the competing-centroid distance margin defined in Eq.~(\ref{eq:competitive_evidence}). A positive margin indicates that the assigned centroid is closer than its strongest competitor, while a larger value reflects clearer inter-cluster separation. As shown in Fig.~\ref{fig:margin_distribution}, CGPLR shifts the distribution toward larger positive margins relative to the clustering baseline, indicating fewer samples near ambiguous cluster boundaries.

The additional computation introduced by CGPLR arises mainly from nearest-neighbor search, candidate-cluster scoring, temporal history storage and the weak--strong consistency branch. Table~\ref{tab:efficiency} quantifies this overhead relative to the clustering baseline.

\begin{table}[!t]
\centering

\caption{Computational comparison of the clustering baseline and
CGPLR on MSMT17$\rightarrow$Market-1501. Adaptation time and peak
memory are reported for 50 adaptation rounds using one NVIDIA RTX
4090.}
\label{tab:efficiency}

\begingroup
\scriptsize
\setlength{\tabcolsep}{2.5pt}
\renewcommand{\arraystretch}{0.92}

\begin{tabular}{@{}lcccc@{}}
\toprule
Method
& \shortstack{Adaptation\\time (h)}
& \shortstack{Time\\ratio ($\times$)}
& \shortstack{Peak\\memory (GB)}
& FPS \\
\midrule

Clustering baseline
& 4.82
& 1.00
& 6.3
& 302 \\

+ Multi-evidence reliability
& 5.35
& 1.11
& 6.7
& 302 \\

+ Selective refinement
& 5.61
& 1.16
& 6.9
& 301 \\

\textbf{Full CGPLR}
& \textbf{6.07}
& \textbf{1.26}
& \textbf{7.1}
& \textbf{300} \\

\bottomrule
\end{tabular}

\vspace{1mm}

\endgroup
\end{table}

Table~\ref{tab:efficiency} evaluates the computational overhead of the reliability-aware adaptation process. CGPLR retains the same ResNet-50 encoder as the clustering baseline and introduces no additional inference-time branch; therefore, the parameter count and inference FLOPs remain unchanged at 23.51M and 4.09G, respectively, with nearly identical retrieval speed. The additional cost is confined to target-domain adaptation, where nearest-neighbor analysis, competing-centroid evaluation, temporal-history tracking, candidate-cluster scoring, and weak--strong consistency learning are performed. Consequently, the full framework requires approximately 1.26$\times$ the adaptation time and 0.8~GB more peak memory than the baseline. This moderate training-only overhead is justified by the retrieval improvements and introduces no additional parameters, FLOPs, or auxiliary modules during inference.

\section{Conclusion}
\label{sec:conclusion}

We introduced CGPLR, a confidence-guided pseudo-label refinement framework for unsupervised cross-domain person re-identification. The framework estimates target-sample reliability using centroid compatibility, neighborhood consistency, competing-centroid separation and temporal assignment stability and makes cluster-adaptive preserve, reassign or defer decisions. Preserved and reassigned samples are optimized through reliability-weighted classification and metric learning, while deferred samples contribute through label-free representation consistency. Experiments on three benchmark datasets, together with ablation, pseudo-label reliability, sensitivity, margin-distribution, qualitative and computational analyses, demonstrate the complementary contributions of the proposed mechanisms across multiple cross-domain adaptation settings. These results show that reliability is useful not only for weighting pseudo-labeled samples but also for determining the form of supervision they receive. Future work will explore more adaptive reliability estimation to reduce dependence on the initial cluster structure and evaluate CGPLR on larger and dynamically changing target domains.

\bibliographystyle{IEEEtran}
\bibliography{References}

@inproceedings{dai2022ccl,
  title     = {Cluster Contrast for Unsupervised Person Re-Identification},
  author    = {Dai, Zuozhuo and Wang, Guangyuan and Yuan, Weihao and Zhu, Siyu and Tan, Ping},
  booktitle = {Proceedings of the Asian Conference on Computer Vision},
  pages     = {1142--1160},
  year      = {2022}
}

@article{tian2024dcl,
  title   = {{DCL}: Dipolar Confidence Learning for Source-Free Unsupervised Domain Adaptation},
  author  = {Tian, Qing and Sun, Heyang and Peng, Shun and Zheng, Yuhui and Wan, Jun and Lei, Zhen},
  journal = {IEEE Transactions on Circuits and Systems for Video Technology},
  volume  = {34},
  number  = {6},
  pages   = {4342--4353},
  year    = {2024},
  doi     = {10.1109/TCSVT.2023.3332353}
}

@article{tan2026ahn,
  title   = {Unsupervised lifelong person re-identification via affinity harmonization},
  author  = {Tan, J. and Peng, J. and Zhang, S. and Wang, Z. and Wang, H.},
  journal = {ACM Transactions on Multimedia Computing, Communications and Applications},
  volume  = {22},
  number  = {4},
  pages   = {1--22},
  year    = {2026}
}

@inproceedings{cho2022pplr,
  title     = {Part-based pseudo label refinement for unsupervised person re-identification},
  author    = {Cho, Yoonki and Kim, Woo Jae and Hong, Seunghoon and Yoon, Sung-Eui},
  booktitle = {Proceedings of the IEEE/CVF Conference on Computer Vision and Pattern Recognition},
  pages     = {7308--7318},
  year      = {2022}
}

@inproceedings{zou2023dcmip,
  title     = {Discrepant and multi-instance proxies for unsupervised person re-identification},
  author    = {Zou, Chang and Chen, Zeqi and Cui, Zhichao and Liu, Yuehu and Zhang, Chi},
  booktitle = {Proceedings of the IEEE/CVF International Conference on Computer Vision},
  pages     = {11058--11068},
  year      = {2023}
}

@article{zhu2024dkdmpl,
  title   = {Dual knowledge distillation on multiview pseudo labels for unsupervised person re-identification},
  author  = {Zhu, Wenjie and Peng, Bo and Yan, Wei Qi},
  journal = {IEEE Transactions on Multimedia},
  volume  = {26},
  pages   = {7359--7371},
  year    = {2024}
}

@inproceedings{peng2025cde,
  title     = {{CDE}-learning: Camera deviation elimination learning for unsupervised person re-identification},
  author    = {Peng, Jinjia and Zhang, Songyu and Wang, Huibing},
  booktitle = {Proceedings of the AAAI Conference on Artificial Intelligence},
  volume    = {39},
  pages     = {6452--6460},
  year      = {2025}
}

@inproceedings{lee2023camera,
  title     = {Camera-driven representation learning for unsupervised domain adaptive person re-identification},
  author    = {Lee, Geon and Lee, Sanghoon and Kim, Dohyung and Shin, Younghoon and Yoon, Yongsang and Ham, Bumsub},
  booktitle = {Proceedings of the IEEE/CVF International Conference on Computer Vision},
  pages     = {11453--11462},
  year      = {2023},
  publisher = {IEEE}
}

@article{pandeeswari2025deep,
  title     = {Deep intelligent technique for person re-identification system in surveillance images},
  author    = {Pandeeswari, Ms R Mallika Alias and Rajakumar, G.},
  journal   = {Pattern Recognition},
  volume    = {162},
  pages     = {111349},
  year      = {2025},
  publisher = {Elsevier}
}

@inproceedings{zheng2015market,
  title     = {Scalable person re-identification: A benchmark},
  author    = {Zheng, Liang and Shen, Liyue and Tian, Lu and Wang, Shengjin and Wang, Jingdong and Tian, Qi},
  booktitle = {Proceedings of the IEEE International Conference on Computer Vision},
  pages     = {1116--1124},
  year      = {2015}
}

@inproceedings{wei2018person,
  title     = {Person transfer {GAN} to bridge domain gap for person re-identification},
  author    = {Wei, Longhui and Zhang, Shiliang and Gao, Wen and Tian, Qi},
  booktitle = {Proceedings of the IEEE Conference on Computer Vision and Pattern Recognition},
  pages     = {79--88},
  year      = {2018}
}

@inproceedings{li2014deepreid,
  title     = {{DeepReID}: Deep filter pairing neural network for person re-identification},
  author    = {Li, Wei and Zhao, Rui and Xiao, Tong and Wang, Xiaogang},
  booktitle = {Proceedings of the IEEE Conference on Computer Vision and Pattern Recognition},
  pages     = {152--159},
  year      = {2014}
}

@article{ran2025camera,
  title     = {Camera-aware graph multi-domain adaptive learning for unsupervised person re-identification},
  author    = {Ran, Zhidan and Lu, Xiaobo and Wei, Xuan and Liu, Wei},
  journal   = {Pattern Recognition},
  volume    = {161},
  pages     = {111217},
  year      = {2025},
  publisher = {Elsevier}
}

@article{peng2026bi,
  author={Peng, Jinjia and Liu, Junyu and Zuo, Xutao and Tao, Zeze and Wang, Huibing},
  title={Bi-Level Inter-Modality Modulation for Unsupervised Visible-Infrared Person Re-Identification},
  journal={IEEE Transactions on Information Forensics and Security},
  year={2026},
  volume={21},
  pages={2652--2667},
  doi={10.1109/TIFS.2026.3671087}
}

@article{peng2024adapt,
  title     = {Adapt only once: Fast unsupervised person re-identification via relevance-aware guidance},
  author    = {Peng, Jinjia and Yu, Jiazuo and Wang, Chengjun and Wang, Huibing and Fu, Xianping},
  journal   = {Pattern Recognition},
  volume    = {150},
  pages     = {110360},
  year      = {2024},
  publisher = {Elsevier}
}

@article{zhang2025ada,
  author={Zhang, Wei and Ye, Peijun and Chen, Dihu and Su, Tao},
  title={ADA Framework for Unsupervised Domain Adaptation Person Re-Identification},
  journal={Pattern Recognition},
  volume={171},
  pages={112238},
  year={2026},
  doi={10.1016/j.patcog.2025.112238}

}

@article{qu2025secureda,
  author={Qu, Xiaofeng and Liu, Li and Zhang, Huaxiang and Zhu, Lei and Nie, Liqiang and Chang, Xiaojun and Li, Fengling},
  title={SecureDA: Privacy-Preserving Source-Free Domain Adaptation for Person Re-Identification},
  journal={IEEE Transactions on Multimedia},
  year={2025},
  volume={27},
  pages={7744--7755},
  doi={10.1109/TMM.2025.3599094}
}

@inproceedings{he2022secret,
  title     = {{SECRET}: Self-consistent pseudo label refinement for unsupervised domain adaptive person re-identification},
  author    = {He, Tao and Shen, Leqi and Guo, Yuchen and Ding, Guiguang and Guo, Zhenhua},
  booktitle = {Proceedings of the AAAI Conference on Artificial Intelligence},
  volume    = {36},
  number    = {1},
  pages     = {879--887},
  year      = {2022}
}

@article{miao2024confidence,
  title     = {Confidence-guided centroids for unsupervised person re-identification},
  author    = {Miao, Yunqi and Deng, Jiankang and Ding, Guiguang and Han, Jungong},
  journal   = {IEEE Transactions on Information Forensics and Security},
  volume    = {19},
  pages     = {6471--6483},
  year      = {2024},
  publisher = {IEEE}
}

@article{iqbal2026cdmmd,
  author={Iqbal, Sundas and Tian, Qing and Saeed, Muhammad and Ahmed, Naeem},
  title={Enhancing Person Re-Identification with Dissimilarity-Based Domain Adaptation and Contrastive Learning},
  journal={Image and Vision Computing},
  volume={171},
  pages={105995},
  year={2026},
  doi={10.1016/j.imavis.2026.105995}
}

@inproceedings{dai2021idm,
  title     = {{IDM}: An intermediate domain module for domain adaptive person re-identification},
  author    = {Dai, Yongxing and Liu, Jun and Sun, Yifan and Tong, Zekun and Zhang, Chi and Duan, Ling-Yu},
  booktitle = {Proceedings of the IEEE/CVF International Conference on Computer Vision},
  pages     = {11864--11874},
  year      = {2021}
}

@inproceedings{hu2022divide,
  title     = {Divide-and-regroup clustering for domain adaptive person re-identification},
  author    = {Hu, Zhengdong and Sun, Yifan and Yang, Yi and Zhou, Jianguang},
  booktitle = {Proceedings of the AAAI Conference on Artificial Intelligence},
  volume    = {36},
  number    = {1},
  pages     = {980--988},
  year      = {2022}
}

@article{chen2023mtnet,
  title     = {{MTNet}: Mutual tri-training network for unsupervised domain adaptation on person re-identification},
  author    = {Chen, Si and Qiu, Liuxiang and Tian, Zimin and Yan, Yan and Wang, Da-Han and Zhu, Shunzhi},
  journal   = {Journal of Visual Communication and Image Representation},
  volume    = {90},
  pages     = {103749},
  year      = {2023},
  publisher = {Elsevier}
}

@article{wang2024attention,
  title     = {Attention-disentangled re-identification network for unsupervised domain adaptive person re-identification},
  author    = {Wang, Lun and Huang, Jiapeng and Huang, Luoqi and Wang, Fei and Gao, Changxin and Li, Jinsheng and Xiao, Fei and Luo, Dapeng},
  journal   = {Knowledge-Based Systems},
  volume    = {304},
  pages     = {112583},
  year      = {2024},
  publisher = {Elsevier}
}

@inproceedings{zheng2021online,
  title     = {Online pseudo label generation by hierarchical cluster dynamics for adaptive person re-identification},
  author    = {Zheng, Yi and Tang, Shixiang and Teng, Guolong and Ge, Yixiao and Liu, Kaijian and Qin, Jing and Qi, Donglian and Chen, Dapeng},
  booktitle = {Proceedings of the IEEE/CVF International Conference on Computer Vision},
  pages     = {8371--8381},
  year      = {2021}
}

@article{tang2026unsupervised,
  title   = {Unsupervised person re-identification via camera-aware multi-level label refinement},
  author  = {Tang, Ning and Fan, Zheyi and Zhu, Yixuan and Zhang, Tingting},
  journal = {Neural Networks},
  volume  = {195},
  pages   = {108292},
  year    = {2026},
  doi     = {10.1016/j.neunet.2025.108292}
}

@article{zhang2024asymmetric,
  title   = {Asymmetric double networks mutual teaching for unsupervised person re-identification},
  author  = {Zhang, Miaohui and Li, Kaifang and Ma, Jianxin and Wang, Xile},
  journal = {Neural Networks},
  volume  = {169},
  pages   = {744--755},
  year    = {2024},
  doi     = {10.1016/j.neunet.2023.11.001}
}

@inproceedings{ge2020self,
  title     = {Self-paced Contrastive Learning with Hybrid Memory for Domain Adaptive Object Re-Identification},
  author    = {Ge, Yixiao and Zhu, Feng and Chen, Dapeng and Zhao, Rui},
  booktitle = {Advances in Neural Information Processing Systems},
  volume    = {33},
  pages     = {11309--11321},
  year      = {2020}
}

@inproceedings{wu2022multi,
  title     = {Multi-Centroid Representation Network for Domain Adaptive Person Re-Identification},
  author    = {Wu, Yuhang and Huang, Tengteng and Yao, Haotian and Zhang, Chi and Shao, Yuanjie and Han, Chuchu and Gao, Changxin and Sang, Nong},
  booktitle = {Proceedings of the AAAI Conference on Artificial Intelligence},
  volume    = {36},
  number    = {3},
  pages     = {2750--2758},
  year      = {2022}
}

@article{zhang2025heterogeneous,
  title   = {Cross-domain person re-identification via learning Heterogeneous Pseudo Labels},
  author  = {Zhang, Zhong and He, Di and Liu, Shuang},
  journal = {Pattern Recognition},
  volume  = {166},
  pages   = {111702},
  year    = {2025},
  doi     = {10.1016/j.patcog.2025.111702}
}

@article{li2026dynamic,
  author  = {Zhi-Hua Li and Xue-Yan Wang and Si-Bao Chen and Chris H. Q. Ding and Bin Luo},
  title   = {Dynamic adaptive multi-view contrastive learning for unsupervised person re-identification},
  journal = {Neural Networks},
  volume  = {203},
  pages   = {109083},
  year    = {2026},
  doi     = {10.1016/j.neunet.2026.109083}
}

@article{chen2026multigranularity,
  author  = {Yanbing Chen and Lingyi Guo and Zhixin Tie and Yinghong Xu and Hao Sheng},
  title   = {Multi-granularity collaborative constraint feature alignment network for unsupervised person re-identification},
  journal = {Pattern Recognition},
  volume  = {179},
  pages   = {113676},
  year    = {2026},
  doi     = {10.1016/j.patcog.2026.113676}
}

@article{zheng2025threec,
  title   = {{3C}: Confidence-guided clustering and contrastive learning for unsupervised person re-identification},
  author  = {Zheng, Mingxiao and Qu, Yanpeng and Li, Dongxuan and Shang, Changjing and Yang, Longzhi and Shen, Qiang},
  journal = {Neurocomputing},
  volume  = {641},
  pages   = {130368},
  year    = {2025},
  doi     = {10.1016/j.neucom.2025.130368}
}

@article{gao2024semantic,
  title     = {A Semantic Perception and CNN-Transformer Hybrid Network for Occluded Person Re-Identification},
  author    = {Gao, Zan and Chen, Peng and Zhuo, Tao and Liu, Meng and Zhu, Lei and Wang, Meng and Chen, Shengyong},
  journal   = {IEEE Transactions on Circuits and Systems for Video Technology},
  volume    = {34},
  number    = {4},
  pages     = {2010--2025},
  year      = {2024},
  doi       = {10.1109/TCSVT.2023.3296680},
  publisher = {IEEE}
}

@article{ding2026multi,
  title     = {Multi-Scale Adaptive Clustering and Local Consistency Learning for Unsupervised Clothing-Changing Person Re-Identification},
  author    = {Ding, Yongkang and Ye, Zi and Xu, Ivonne and Lyu, Shuangquan and Zhang, Liyan},
  journal   = {IEEE Transactions on Information Forensics and Security},
  volume    = {21},
  pages     = {2889--2904},
  year      = {2026},
  doi       = {10.1109/TIFS.2026.3671089},
  publisher = {IEEE}
}

@article{yang2026attention,
 author  = {Zhiwei Yang and Jing Liu and Guansong Pang and Peng Wu and Zhaoyang Wu},
  title   = {Attention-driven pseudo-label self-training for weakly supervised video anomaly detection},
  journal = {Pattern Recognition},
  volume  = {177},
  pages   = {113349},
  year    = {2026},
  doi     = {10.1016/j.patcog.2026.113349}
}

@inproceedings{ester1996density,
  author    = {Ester, Martin and Kriegel, Hans-Peter and Sander, J{\"o}rg and Xu, Xiaowei},
  title     = {A Density-Based Algorithm for Discovering Clusters in Large Spatial Databases with Noise},
  booktitle = {Proceedings of the Second International Conference on Knowledge Discovery and Data Mining},
  year      = {1996},
  pages     = {226--231},
  publisher = {AAAI Press}
}

@inproceedings{zhong2017reranking,
  author    = {Zhong, Zhun and Zheng, Liang and Cao, Donglin and Li, Shaozi},
  title     = {Re-Ranking Person Re-Identification with $k$-Reciprocal Encoding},
  booktitle = {Proceedings of the IEEE Conference on Computer Vision and Pattern Recognition},
  year      = {2017},
  pages     = {1318--1327}
}

@article{tian2025unsupervised,
  author={Tian, Qing and Liu, Xiang and Sun, Jixin and Wan, Jun and Lei, Zhen},
  title={Unsupervised Domain Adaptation Person Re-Identification: Bridged by Feature Fusion Transitional Domain},
  journal={IEEE Transactions on Information Forensics and Security},
  year={2025},
  volume={20},
  pages={11417--11430},
  doi={10.1109/TIFS.2025.3607258}
}

@article{tian2025cross,
  title={Cross-attention with conditional matching for multi-target domain adaptation},
  author={Tian, Qing and Liu, Xiang and Zhou, Jiazhong and Zheng, Yuhui and Wan, Jun and Lei, Zhen},
  journal={IEEE Transactions on Circuits and Systems for Video Technology},
  volume={35},
  number={11},
  pages={10918--10929},
  year={2025},
  publisher={IEEE}
}

@article{tian2026progressive,
  author={Tian, Qing and Shen, Junyu and Kang, Lulu and Ou, Weihua and Wan, Jun and Lei, Zhen},
  title={Progressive Curriculum Learning With Teacher-Student Collaboration for Source-Free Unsupervised Domain Adaptation},
  journal={IEEE Transactions on Circuits and Systems for Video Technology},
  year={2026},
  volume={36},
  number={2},
  pages={1627--1639},
  doi={10.1109/TCSVT.2025.3601438}
}

@article{tian2025duel,
  author  = {Qing Tian and Junjie Pan and Yun Yang and Weihua Ou},
  title   = {Dual-focus memory contrastive learning for active domain adaptation},
  journal = {Neural Networks},
  volume  = {195},
  pages   = {108224},
  year    = {2026},
  issn    = {0893-6080},
  doi     = {10.1016/j.neunet.2025.108224},
}

@article{qin2026oneshot,
  author  = {Qin, Jinghui and Tan, Hao and Jin, Kebing},
  title   = {One-shot Unsupervised Domain Adaptation for Semantic Segmentation with Self-Constrained Style Decoupling},
  journal = {IEEE Transactions on Circuits and Systems for Video Technology},
  year    = {2026},
  note    = {{Early Access}, doi: 10.1109/TCSVT.2026.3702681}
}

@article{wang2026temporal,
  author={Wang, Yucheng and Gong, Peiliang and Wu, Min and Ott, Felix and Li, Xiaoli and Xie, Lihua and Chen, Zhenghua},
  title={Temporal Source Recovery for Time-Series Source-Free Unsupervised Domain Adaptation},
  journal={IEEE Transactions on Pattern Analysis and Machine Intelligence},
  year={2026},
  note={{Early Access}, doi: 10.1109/TPAMI.2026.3681762}
}

@article{yang2026multilevel,
  author={Yang, Xi and Zhou, Hexun and Cheng, De and Tian, Menghui and Wang, Nannan},
  title={Multi-level Explicit Feature Alignment Network for Unsupervised Domain Adaptive Person Search},
  journal={IEEE Transactions on Circuits and Systems for Video Technology},
  year={2026},
  note={{Early Access}, doi: 10.1109/TCSVT.2026.3704656}
}

@article{tian2025part,
  author={Tian, Qing and Wang, Bin and Liu, Xiang and Shen, Jiashuo and Cheng, Keyang and Ou, Weihua and Lei, Zhen},
  journal={IEEE Transactions on Circuits and Systems for Video Technology},
  title={Part-Based Feature Complementary Denoising for Unsupervised Person Re-Identification},
  year={2026},
  volume={36},
  number={2},
  pages={2137--2150},
  doi={10.1109/TCSVT.2025.3609570}
}

@article{wang2026structured,
  author  = {Wang, Yubin and Yu, Huimin and Hu, Haoji and Gao, Yongsheng},
  title   = {Structured Distilled 3D Gait Fields for Clothes-Changing Person Re-Identification in Videos},
  journal = {IEEE Transactions on Circuits and Systems for Video Technology},
  year    = {2026},
  note    = {{Early Access}, doi: 10.1109/TCSVT.2026.3691145}
}

@ARTICLE{Yang2026MEFAN,
  author  = {Yang, Xi and Zhou, Hexun and Cheng, De and Tian, Menghui and Wang, Nannan},
  title   = {Multi-level Explicit Feature Alignment Network for Unsupervised Domain Adaptive Person Search},
  journal = {IEEE Transactions on Circuits and Systems for Video Technology},
  year    = {2026},
  pages   = {1--1},
  doi     = {10.1109/TCSVT.2026.3704656}
}

@ARTICLE{Liu2026Confidence,
  author  = {Liu, Yang and Zhang, Jiahua and Wu, Yue and Zhao, Zijing and Chen, Qingchao and Peng, Yuxin},
  title   = {Confidence-Aware Pseudo-Label Self-Correction for Weakly Supervised Visual Grounding},
  journal = {IEEE Transactions on Pattern Analysis and Machine Intelligence},
  year    = {2026},
  volume  = {48},
  number  = {8},
  pages   = {8901--8918},
  doi     = {10.1109/TPAMI.2026.3673741}
}

@ARTICLE{Zhang2026OpenBPR,
  author  = {Zhang, Guanjia and Xing, Weiwei and Liu, Weibin and Sang, Fusong and Wang, Xinyue and Wei, Xiang},
  title   = {OpenBPR: Bias-Guided Pseudo-Label Refinement for Open-World Semi-Supervised Learning},
  journal = {IEEE Transactions on Circuits and Systems for Video Technology},
  year    = {2026},
  volume  = {36},
  number  = {8},
  pages   = {11858--11875},
  doi     = {10.1109/TCSVT.2026.3685347}
}

\begin{IEEEbiography}
[{\includegraphics[width=1in,height=1.25in,clip,keepaspectratio]{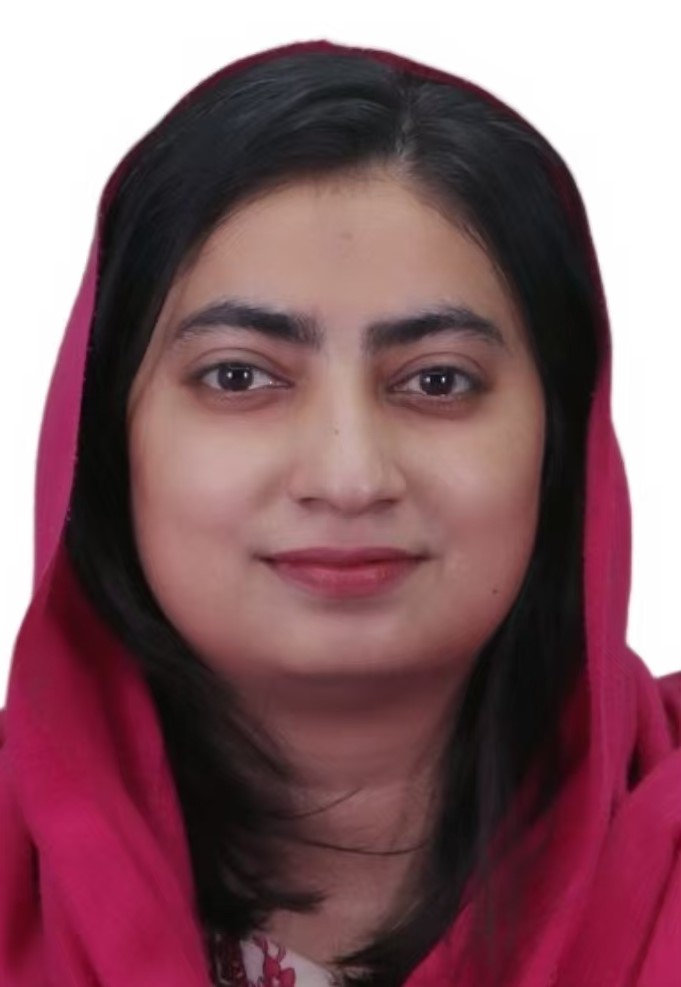}}]
{Sundas Iqbal}
received her M.S. degree in Computer Science from Pakistan. She is currently a Ph.D. scholar at Nanjing University of Information Science and Technology (NUIST), China. Her research focuses on machine learning, domain adaptation and domain generalization. Her research interests include cross-domain visual recognition, person re-identification and robust learning under distribution shift.
\end{IEEEbiography}

\vspace{-12pt}

\begin{IEEEbiography}
[{\includegraphics[width=1in,height=1.25in,clip,keepaspectratio]{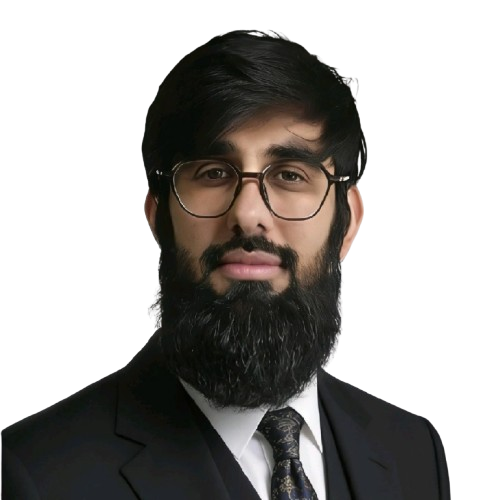}}]
{Danish Ali}
received his B.S. degree in Computer Science from the University of Haripur, Pakistan and his M.S. degree in Software Engineering from Wuhan University, China, in 2026. He is currently pursuing the Ph.D. degree at Wuhan University, China. His research interests include domain adaptation, person re-identification, deep learning, computer vision, natural language processing, large language models and intelligent systems.
\end{IEEEbiography}

\begin{IEEEbiography}
[{\includegraphics[width=1in,height=1.25in,clip,keepaspectratio]{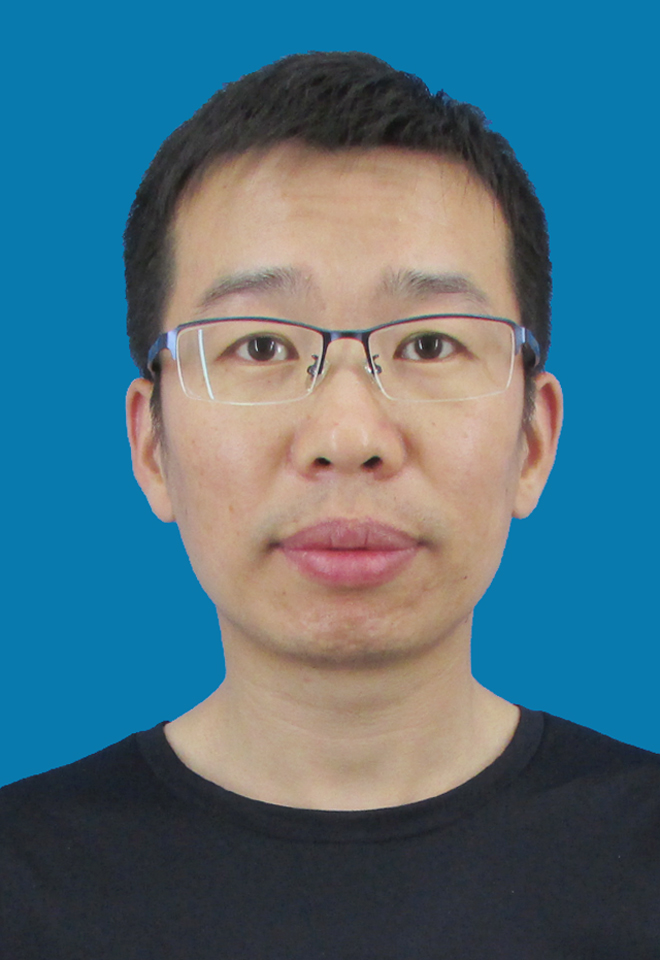}}]
{Qing Tian}
received his Ph.D. degree in computer science from Nanjing University of Aeronautics and Astronautics, China, in 2016. He is currently a Professor in the School of Software at Nanjing University of Information Science and Technology, China. He was an Academic Visitor at the University of Manchester, UK from 2018 to 2019. He is the recipient of the National PhD Scholarship Award of China in 2015, the Best Scientific Paper Award of ICPR in 2016 and the Excellent Doctoral Dissertation Award of Jiangsu Province of China in 2017. He has published over thirty peer-reviewed scientific articles and served as a program committee member for several renowned international conferences such as AAAI, IJCAI, CVPR, ICPR, PAKDD and as a reviewer for many prestigious journals and conferences including IEEE TPAMI, IEEE TKDE, IEEE TNNLS, IEEE TCSVT, IEEE TCYB, IEEE TIFS, AAAI, IJCAI, ICDM and CVPR. His research interests include machine learning, pattern recognition and computer vision.
\end{IEEEbiography}

\begin{IEEEbiography}
[{\includegraphics[width=1in,height=1.25in,clip,keepaspectratio]{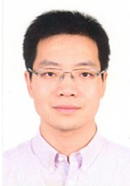}}]
{Yuhui Zheng~(Member, IEEE)}
was born in Shaanxi, China, in 1982. He received the B.Sc. degree in pharmacy engineering and the Ph.D. degree in pattern recognition and intelligent system from Nanjing University of Science and Technology, Nanjing, China, in 2004 and 2009, respectively. From 2014 to 2015, he was a Visiting Professor with the Digital Media Laboratory, School of Electronic and Electrical Engineering, Sungkyunkwan University, South Korea. He is currently a Full Professor at the College of Computer, Qinghai Normal University. His main research interests include image and video analysis, scene understanding, visual tracking and pattern recognition.
\end{IEEEbiography}

\begin{IEEEbiography}
[{\includegraphics[width=1in,height=1.25in,clip,keepaspectratio]{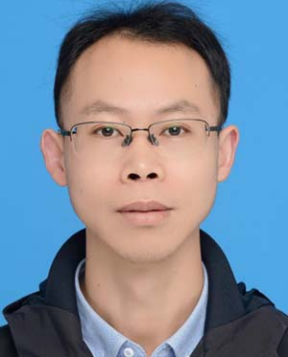}}]
{Jianping Gou~(Senior Member, IEEE)} received the PhD degree in computer science from the University of Electronic Science and Technology of China, Chengdu, China, in 2012. He was a post-doctoral research fellow with the University of Sydney. He is currently a professor with the College of Computer and Information Science, Southwest University, Chongqing, China. His current research interests include pattern classification and machine learning. So far, he has published more than 160 papers in international journals and conference proceedings, such as IJCV and CVPR.
\end{IEEEbiography}

\begin{IEEEbiography}
[{\includegraphics[width=1in,height=1.25in,clip,keepaspectratio]{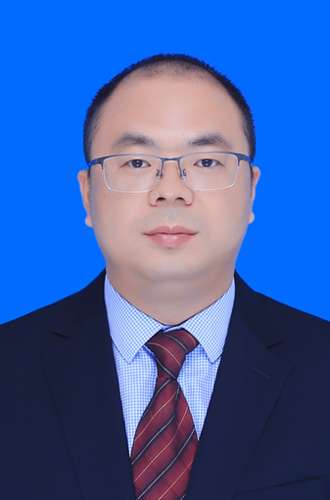}}]
{Weihua Ou}
received the Ph.D. degree in information and communication engineering from the Huazhong University of Science and Technology, Wuhan, China, in 2014. From 2016 to 2017, he was a Post-Doctoral Researcher with the University of Technology Sydney, NSW, Australia. He has authored or co-authored more than 60 academic papers in top journals and conferences including IEEE Transactions on Multimedia, IEEE Transactions on Neural Networks and Learning Systems, and Pattern Recognition. His research interests include computer vision and pattern recognition. He served as a Program Committee Member of top conferences, such as IJCAI, AAAI and CVPR.
\end{IEEEbiography}

\end{document}